\documentclass[pdflatex,sn-mathphys-num]{sn-jnl}% Math and Physical Sciences Numbered Reference Style 
\usepackage{graphicx}%
\usepackage[T1]{fontenc}
\usepackage{multirow}%
\usepackage{amsmath,amssymb,amsfonts}%
\usepackage{amsthm}%
\usepackage{mathrsfs}%
\usepackage[title]{appendix}%
\usepackage{xcolor}%
\usepackage{textcomp}%
\usepackage{manyfoot}%
\usepackage{booktabs}%
\usepackage{algorithm}%
\usepackage{algorithmicx}%
\usepackage{algpseudocode}%
\usepackage{listings}%
\usepackage{hyperref}
\usepackage[outline]{contour}% http://ctan.org/pkg/contour

\usepackage[numbers]{natbib}%
\usepackage{multirow}
\usepackage{comment}

\theoremstyle{plain}
\begin{document}

\title[Article Title]{A Review on Geometry and Surface Inspection in 3D Concrete Printing
}

% title variations
\begin{comment}
A Review on Optical Inspection of Geometry and Surface in 3D Concrete Printing

Aspects of geometrical  quality control in digital construction  : A review

A Review of Geometric and surface Quality Control in 3D Concrete Printing: Key Aspects and Influencing Factors

A Review on Geometry and Surface Inspection in 3D Concrete Printing

Review paper: Quality Inspection Aspects

Aspects of Quality Control in digital construction: A review

Quality Control in digital construction: A review

Geometric Quality Control in Digital Construction: A Review

Geometric Quality Control in digital construction: A review & Integration

Geometric Quality Control in digital construction: a holistic review
\end{comment}

\begin{comment}
Journal options

1. Cement & Concrete Research
2. ISPRS Journal
3. Automation in construction (this one for filament 2D)
4. Additive Manufacturing (https://www.sciencedirect.com/journal/additive-manufacturing)
\end{comment}


\author*[1]{\fnm{Karam} \sur{Mawas}}\email{k.mawas@tu-bs.de}

\author[1]{\fnm{Mehdi} \sur{Maboudi}}\email{m.maboudi@tu-bs.de}

\author[1]{\fnm{Markus} \sur{Gerke}}\email{m.gerke@tu-bs.de}

\affil[1]{\orgdiv{Institute of Geodesy and Photogrammetry}, \orgname{Technische Universit\"at Braunschweig}, \orgaddress{\street{Bienroder Weg 81}, \city{Braunschweig}, \postcode{38106}, \country{Germany}}}

%\affil[2]{\orgdiv{Department}, \orgname{Organization}, \orgaddress{\street{Street}, \city{City}, \postcode{10587}, \state{State}, \country{Country}}}

%\affil[3]{\orgdiv{Department}, \orgname{Organization}, \orgaddress{\street{Street}, \city{City}, \postcode{610101}, \state{State}, \country{Country}}}


\abstract{
Given the substantial growth in the use of additive manufacturing in construction (AMC), it is necessary to ensure the quality of printed specimens which can be much more complex than conventionally manufactured parts. This study explores the various aspects of geometry and surface quality control for 3D concrete printing (3DCP), with a particular emphasis on deposition-based methods, namely extrusion and shotcrete 3D printing (SC3DP). A comprehensive overview of existing quality control (QC) methods and strategies is provided and preceded by an in-depth discussion. Four categories of data capture technologies are investigated and their advantages and limitations in the context of AMC are discussed. Additionally, the effects of environmental conditions and objects' properties on data capture are also analyzed. The study extends to automated data capture planning methods for different sensors. Furthermore, various quality control strategies are explored across different stages of the fabrication cycle of the printed object including: (i) During printing, (ii) Layer-wise, (iii) Pre-assembly, and (iv) Assembly. In addition to reviewing the methods already applied in AMC, we also address various research gaps and future trends and highlight potential methodologies from adjacent domains that could be transferred to AMC.
}




\maketitle

\section{Introduction}\label{Intro_sec1}

The traditional method of concrete construction involves the pouring of material into molds. However, the emergence of additive manufacturing in construction (AMC) has revolutionized this process by eliminating the need for casts. During the past decade, researchers have investigated AMC, demonstrating its potential for productivity gains and a reduction in labor costs \cite{XU2020,Placzek2021,Senthilnathan2022}. To meet the digitization requirements, which are inherent to AMC, the concept of the digital twin (DT) is fundamental. This is due to its distinctive attribute of facilitating a bidirectional data flow between physical and digital domains. The interested reader can refer to \cite{BERGS2021}, for comprehensive insights into this concept.

\begin{figure}[h]
\centering
\includegraphics[width=0.9\textwidth]{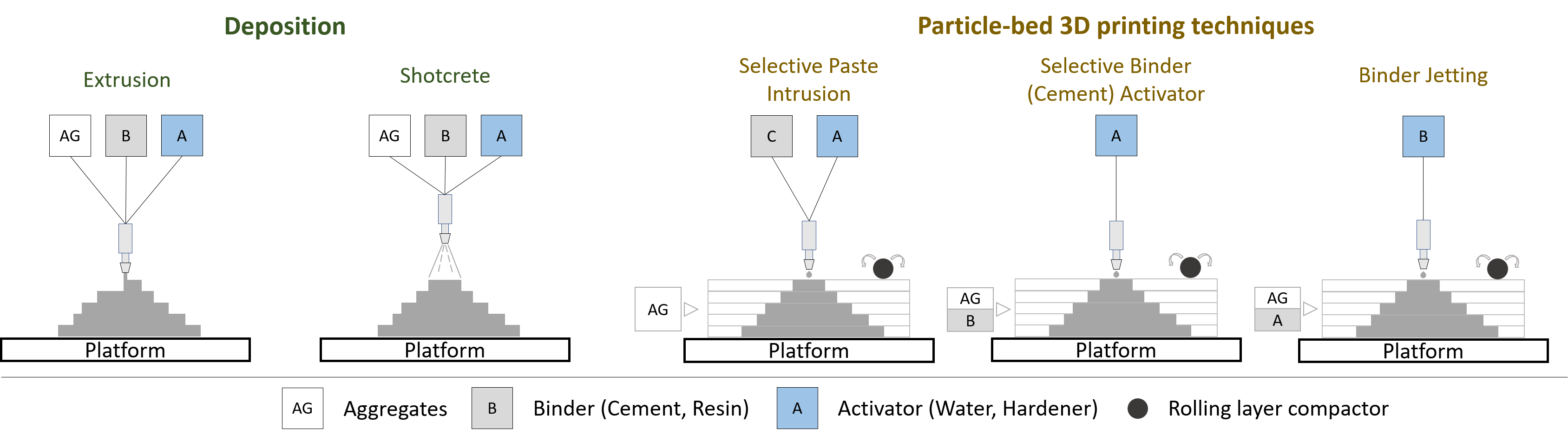}
\caption{Schematic of deposition and particle-bed 3D printing, adopted from \cite{LOWKE2018,Kloft2019}.}\label{Deposition_and_ParticleBed}
\end{figure}

Additive manufacturing (AM) encompasses a multitude of processes, as illustrated in Fig.~\ref{Deposition_and_ParticleBed}. The specific printing strategy employed is determined by the scale of the object being printed and the type of robotic technology utilized. The printing of objects is carried out using a variety of techniques, from two-dimensional printing, such as particle bed printing (at the small-scale level), to 3D printing of objects, including methods such as injection 3D printing  (at the small-scale level) and material extrusion - deposition (at the large-scale level). Various fixed and mobile robotic systems are used in AMC \cite{ArnaudandAmziane2019}, such as the articulated fixed robotic arm \cite{xtreeE}, gantry system \cite{COBOD}, cable robots \cite{Bosscher2005,Izard2017}, and mobile robots \cite{Avenco}. For further information on AM classification, the interested reader might refer to \cite{Wangler2016,DILBEROGLU2017,ArnaudandAmziane2019,Hack2020,Placzek2023}.

\begin{figure}[h!]
    \centering
    \begin{tabular}{cc}
        \includegraphics[width=0.4\textwidth, height=0.2\textheight]{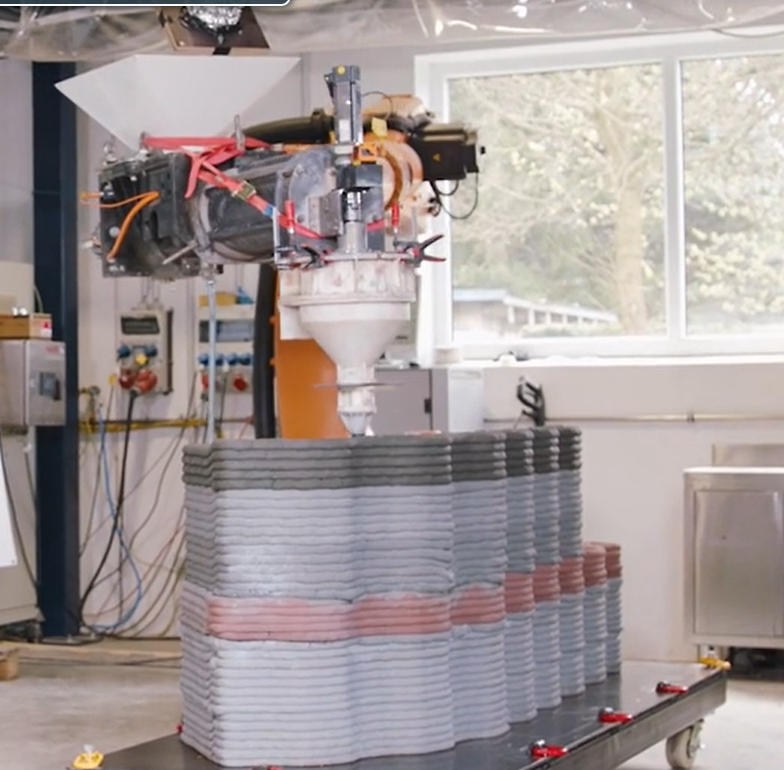}
        % width=0.087
        & 
        \includegraphics[width=.4\textwidth, height=0.2\textheight]{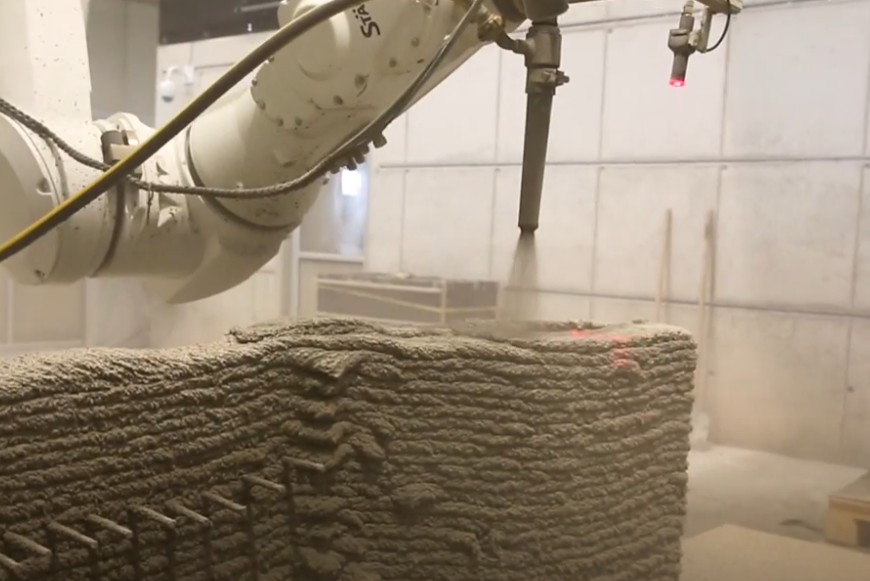} 
        %width=0.038
        \\
        a & b \\
    \end{tabular}
    \caption{Empirical examples of 3D concrete printing; a) Extrusion Concrete printing, b)  Shotcrete 3D printing, images credit: \cite{AMC-TRR277}}
    \label{fig:Real_picture_4_Extrusion_SC3DP}
\end{figure}

In order to limit the scope of this paper, we will focus on the two common large-scale deposition techniques extrusion and shotcrete for 3D concrete printing (3DCP), see Fig.~\ref{fig:Real_picture_4_Extrusion_SC3DP}, which result in the deposition of concrete in long filaments \cite{Dressler2020}. In general, extrusion involves extruding material layer by layer with a controlled nozzle, while SC3DP involves spraying material, resulting in the construction of concrete layers \cite{KLOFT2020}. Compared to extrusion, SC3DP has the potential to overcome the risk of cold joints by improving the interlock between adjacent filaments \cite{KLOFT2020}. In addition, SC3DP offers more geometric freedom compared to extrusion \cite{Heidarnezhad2022}. 

In addition to the printing techniques, three distinct categories, namely \textit{in situ, on-site}, and \textit{off-site} concrete printing, represent a range of conceptual approaches within the AMC. In situ printing requires the utilization of an accurately placed printing robot to ensure that the object is printed at the designed location accurately. In contrast, on-site AMC is distinguished by the production of components in close proximity to their intended location, thereby necessitating the transportation or placement of the fabricated components to their final position. Off-site AMC exhibits similarities to conventional precast production, where components are manufactured in a laboratory environment \cite{Placzek2021_German}.

Quality control (QC) for geometric and textural analysis is crucial to ensure that the fabricated components replicate the design specifications and that pre-defined tolerances are kept. Conventional QC techniques and strategies in construction are inefficient, costly, and less flexible. This is due to the fact that conventional surveying methodologies, such as tacheometry, are implemented at the point-level. However, this approach does not meet the requirements coming with the possibilities of the mentioned AM techniques. The rapid emergence of AMC and their diverse conceptualizations in construction necessitates the adaptation of advanced QC strategies \cite{Placzek2021}. The utilization of digital tools, customized inspection techniques and the deployment of appropriate sensors ensures the uninterrupted QC to quantify deviations from design specifications \cite{Mawas2022_Automatic-Inspection, Slepicka2024,OTTO2024}.

\begin{figure}[t]
\centering
\includegraphics[width=0.55\textwidth]{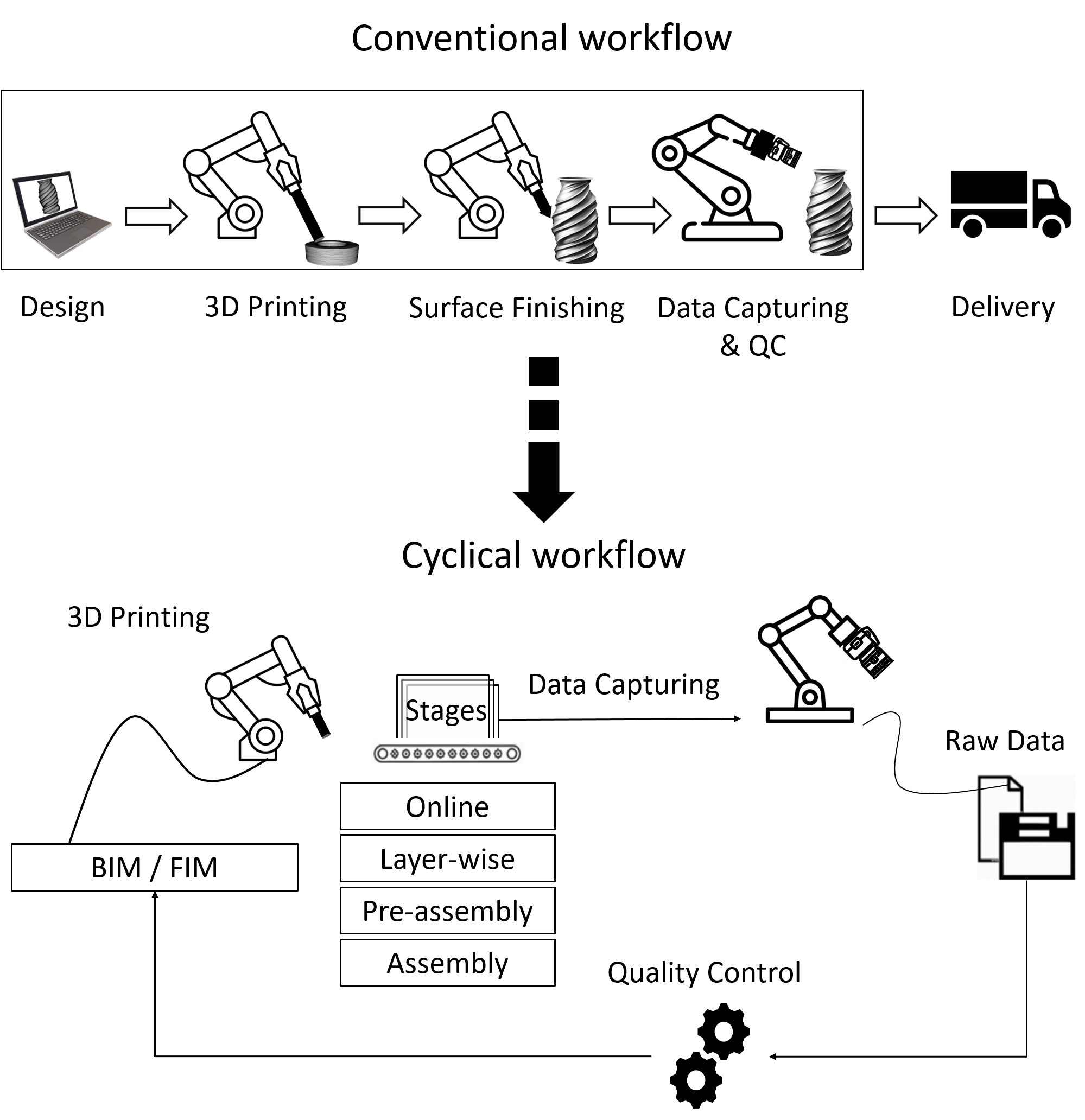}
\caption{Comparison of cyclical workflow for comprehensive QC of 3D printed elements and traditional sequential QC}\label{MY_QC_Diagram}
\end{figure}

The printed objects undergo a variety of different processes and parameters throughout their production. Thereby, there is a need for quality control measures at various stages to ensure the desired outcome \cite{Nair2022}. Stage-wise classification is beneficial for identifying the appropriate application of QC within the fabrication process workflow, as it facilitates the identification of the specific stages where QC measures are both relevant and necessary. As illustrated in Fig.~\ref{MY_QC_Diagram}, there is a transition from a conventional QC (sequential workflow) to a cyclical QC workflow, with multiple stages occurring during the printing phases. The integration of stage-wise QC ensures that printed objects undergo continuous inspection with bi-directional communication and updates between digital models --such as building/fabrication information model (BIM/FIM)-- and physical objects \cite{WOLFS2024}.

The cyclic approach offers distinct advantages over conventional methods. Conventional QC usually takes place at the final stage of printing, mainly before delivery, which limits the ability to make changes and fix defects. This might lead to rejection of the printed component which is expensive and inefficient.
\begin{comment}
    \mg{Disadvantages? Address may be both in conclusions/discussions?}
    KM:  I think what we have now is enough could you check again. from tacheometer and why stage-wise. also check the conclusion
\end{comment}

By implementing continuous stage-wise QC at various printing processes, corrections or adaptations can be made to the printed specimen or designed model in the digital space, ensuring consistency and precision throughout the production process. To this end, the right sensor should be chosen for the task, and dedicated methods for checking the geometric quality of the printed object and surface texture at different stages of the printing process should be employed. This can be realized within a framework that allows for seamless integration and bi-directional data workflow and communication within the different sensors and robots.

\begin{figure}[h]
\centering
\includegraphics[width=0.55\textwidth]{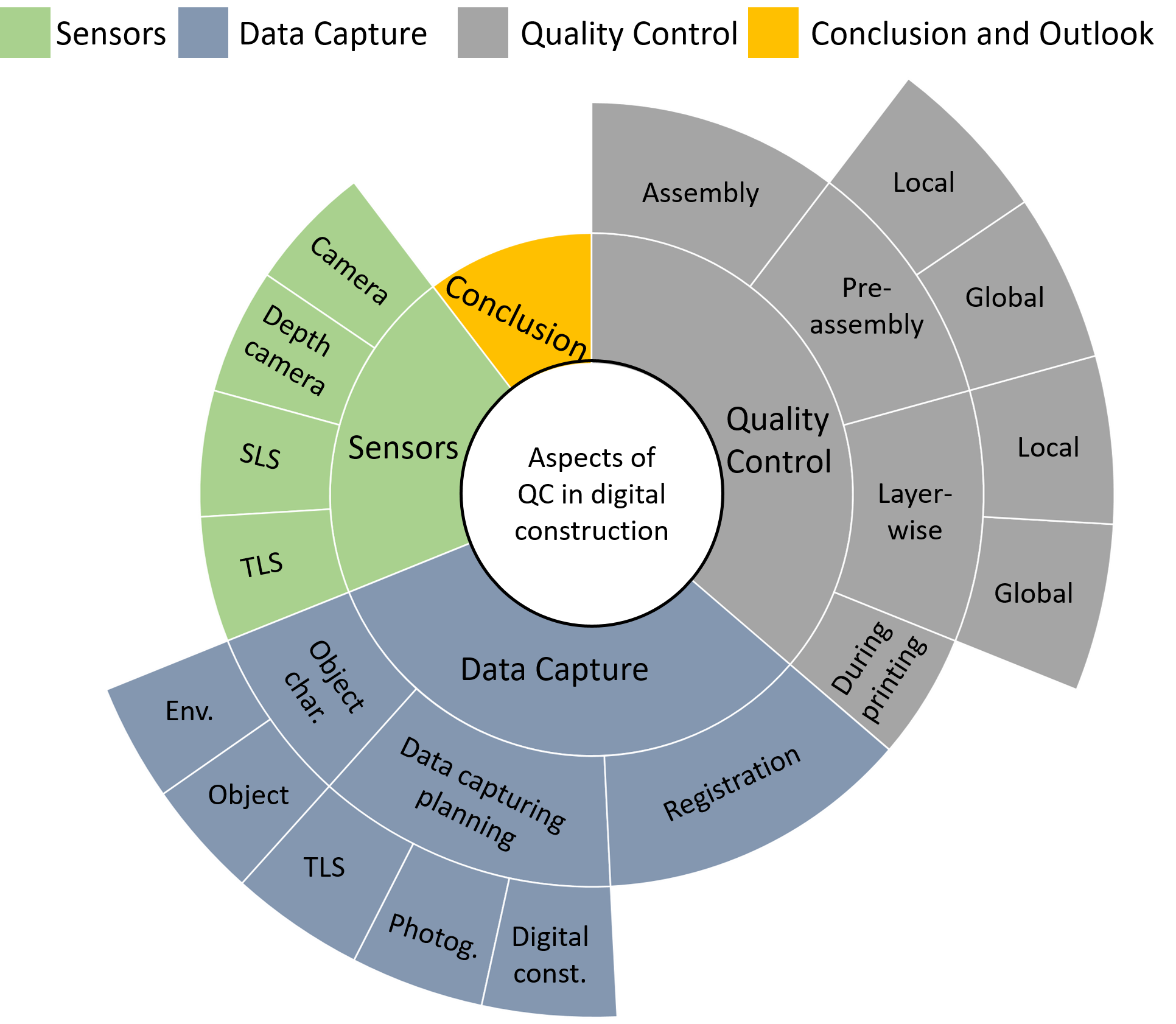}
\caption{Structure of the paper and its key aspects.}\label{Paper_Structure}
\end{figure}

This paper provides a comprehensive investigation of the various sensors employed for reality capture and the techniques for quality inspection in AMC, with a particular focus on 3DCP utilizing both extrusion and shotcreting. The discussed sensors in this paper include RGB cameras, depth cameras, structured light systems, and terrestrial laser scanners, selected for their potential in QC applications related to geometry and surface aspects rather than material or mechanical properties. The environmental and object constraints that can affect data capturing and QC processes are analyzed. This provides insight into critical parameters that should be considered while selecting and employing the sensors in different scenarios. In addition, various methods for data capture planning and registration are discussed that aim at increasing the efficiency, accuracy and automation of data acquisition.

The structure of this paper is illustrated in Fig.~\ref{Paper_Structure}.
\begin{comment}
%, which provides an overview of different aspects that are discussed in the next sections. 
\end{comment} 
 Section~\ref{Sensors_sec2} dives into a variety of available sensors and their specifications. Then, section~\ref{sec3: Data Capturing} explores data capture methodologies, with a focus on the environment and the objects’ properties. Additionally, the data capturing planning and registration methods are addressed. Section~\ref{sec4: Quality Control} presents our classification and detailed review of stage-wise QC methods in 3DCP. Finally, section~\ref{sec5: Conclusion} provides a comprehensive discussion, conclusion, and outlook on the topic.

\section{Sensors}\label{Sensors_sec2}
\begin{comment}
    Previously, in sec.~\ref{Intro_sec1}, we \textbf{explored stage-wise QC diagrams} \mg{not clear - you just showed a diagram but not more details; This first sentence is not clear. Can be shortened/removed?!} that can be utilized throughout the printing processes, which are essential to ensure that the as-built specimen replicates the designed model.
\end{comment}
As illustrated previously, QC is essential to ensure that the as-built specimen replicates the designed model. This objective hinges on the availability of comprehensive captured data from the printed object \cite{OMAR2016,Poux2020}. 
%which can be obtained through various reality capturing methods. 
This section investigates the use of optical non-destructive sensors, employing different sensors for data capture of the printed object. QC involves the utilization of different active or passive sensors. The data used are either images or other types of information like 3D point clouds, which can be produced by active sensors like Terrestrial Laser Scanner (TLS),Structured Light Scanner (SLS), and depth cameras, or passive sensors as conventional cameras with the help of employing a photogrammetric method for 3D scene reconstruction. These data capturing sensors play an essential role in controlling the seamless compatibility between the final construction project and its designed counterpart. In this section, we investigate not only the significance of the sensing technologies but also emphasize their association in the context of AMC.

Beside this intuitive objective which is directly related to reality capture, fulfilling the application-dependent criteria, namely Level of Accuracy (LOA), Level of Details (LOD), and Level of Completeness (LOC) should be also considered while data capturing \cite{ARYAN2021}. The term LOD is used to describe the smallest size of an object that can be recognized \cite{Bryan2009,Sanhudo2020}. LOA represents the acceptable tolerance for the positioning accuracy of each individual point within the point cloud \cite{ARYAN2021, Wujanz_2016}. LOC refers to the portion of the surface of a captured object in relation to the entire scan \cite{Ahn2016, ARYAN2021}.

\subsection{Conventional camera }\label{Photogrammetry}
The emergence of reasonably priced, high-quality cameras has accelerated the development of computer vision and photogrammetry-based reality capturing methods and applications, e.g. in the context of component QC \cite{Golparvar-Fard2015,SAFA2015,OMAR2016,RODRIGUEZMARTIN2016,Maboudi-GSW2025}. The photogrammetry workflow is initiated by the structure-from-motion (SfM) method \cite{SfM_Pollefeys2004}, which utilizes a series of images captured from multiple perspectives. This is typically followed by a bundle block adjustment to ensure accuracy. The process later involves dense image matching and three-dimensional reconstruction of the object of interest, employing the principles of multi-view stereo (MVS) or Semi-Global Matching (SGM) \cite{PMVS_Furukawa, Hirschmuller2011_SGM}. Based on the configuration of the image capturing and reference data in object space, photogrammetry offers an adaptive scale and accuracy of reconstruction. Moreover, texture information and geometry of the edges are inherently captured and can be used in further analysis. However, data collection is often manual and requires onsite considerations. It is also sensitive to varying lighting conditions, shadow effects, low-textured surfaces, and occlusions \cite{HAMLEDARI201778}.

Moreover, capturing each structural part of a component from multiple viewpoints, at least two, is essential. Furthermore, these methods are passive, mostly depending on lighting, texture properties, and the constellation of the image block and relative geometry of images w.r.t.\ the object. The scene can only be relatively reconstructed from 2D images by a photogrammetric approach. Thus, the scale of the model is unknown. External information, such as ground control points, scale-bars \cite{Maboudi2021} or calibrated stereo rigs are required for accurate scale determination.

\begin{comment}
    Moreover, as images can recreate 3D scenes relatively\mg{also here:no: images cannot do anything... but say that: the scene can only be reconstructed up to an unknown scale. In stereo rigs or depth cameras with a small baseline this is given, but for large scenes still external info is needed, mention this also below?!}, there are still difficulties in determining the model's proper scale.
\end{comment}

Camera parameters (such as focal length, sensor size,  etc.), relative pose of the cameras (relative distance and angle), object properties (like texture and shape), environmental conditions (e.g. lighting) and object space information (scale bars, GCPs, GNSS info) are the most important parameters that affect the quality of photogrammetry output \citep[Chapters 12 and 13]{Forstner2016} and \citep[Chapters 1 and 7]{Luhmann2020}.
%[\citenum{Forstner2016}, Chapters 12 and 13; \citenum{Luhmann2020}, Chapters 1 and 7].
%\citep[Chapters 12 and 13]{Forstner2016};
%\citep[Chapters 1 and 7]{Luhmann2020}.
%\citep[Chapters 12 and 13][Chapters 1 and 7]{Forstner2016,Luhmann2020};
%\cite{Forstner2016,Luhmann2020}
%(Forstner and Wrobel 2016, Chapters 12 and 13; Luhmann et al. 2020, 3rd edition, Chapters 1 and 7).%
In addition to point clouds, accurate image-based measurement in photogrammetry provides high-resolution details of the printed object with RGB information, enabling texture analysis and accurate 3D reconstruction at multiple scales.

\subsection{Depth camera}\label{Depth camera}
Depth cameras are devices used to capture 2.5D information of an object. The image is usually generated by using a variety of techniques, namely: Light-field/plenoptic cameras or Time-of-Flight (ToF) cameras.
Plenoptic cameras, on one hand, take advantage of a microlensing array between the lens and image sensor plane \cite{Eckstein2021}. The target object is captured by a number of these microlenses based on its distance (farther objects are recorded by fewer microlenses). The resulting image comprises a multifacet pattern. Furthermore, the depth measurements from the shift of these multifacet patterns correspond to different viewing angles. 

On the other hand, a ToF camera emits infrared light and measures the time-of-flight. To achieve this measurement, there are two operational principles: pulsed light and continuous wave amplitude modulation. The first method is limited by a very short time measurement interval, spanning only a few centimeters in depth
\citep[p.~254]{Luhmann2020}.
%(Luhmann et al. 2020, 3rd edition, pp. 254). 
ToF addresses this limitation by measuring the phase shift differences between the emitted and received light. Thus, the distance is computed for every pixel by applying the phase shift method. However, the system might suffer from ambiguities due to the occurrence probability of multiples of the modulation wavelength \cite{Langmann2012,Li2014}. The depth map for the full field of vision may be created since every pixel encodes the distance to its corresponding point in the picture \cite{Maru2021}. Thus, in contrary to photogrammetry, the scale for depth cameras is determined directly.

In terms of accuracy and range, ToF cameras are the most suited to be used in controlled indoor environments as well as for small objects \cite{CONDOTTA2020}. As a result,  ToF cameras are preferable for applications like indoor prefabrication in AMC. However, their use might be affected by practical limitations including: sensitivity to temperature changes, mixed-pixels, or multipath effects, among others \cite{Horaud2016}. Multipath effect occurs when the light reflects off from several surfaces and converges at the same pixel. Whereas, mixed-pixels happen when the reflected beam hits several surfaces at different distances \cite{Frangez2021}. The same phenomenon can be observed for laser scanners (cf. sec.~\ref{TLS}). To reach the optimal performance, internal warm-up of the camera as well as a proper distance to object have to be considered as shown in \cite{Frangez2020}. These distinctions are critical in the 3DCP environment and construction sites, highlighting the importance of choosing the right tool for specific settings and purposes in 3D imaging and quality assessment.

\subsection{Structured light scanner (SLS)}\label{SLS}
A calibrated projector-camera(s) combination is employed to project a variety of coded-strip patterns onto the surface of the object of interest \cite{RiekeZappRoyo2017,ZHANG2018}. Visible light of different widths of patterns per frame -- to cope with any potential ambiguity -- is projected onto the scene and captured by the camera(s).  Additionally, varying the pattern width helps in distinguishing between different areas and reducing errors in depth measurement. The system's capacity to precisely interpret these pixels and decode the projected patterns is attributed to the uniqueness of the coded patterns and their width variations. Furthermore, the system's calibration enables the use of spatial triangulation for every captured pixel, facilitating the estimation of the 3D coordinate of the pixels. Due to this setup, the scale is determined directly, in contrast to 2D cameras with photogrammetric approach for 3D scene reconstruction.
The capture of the whole scene is achieved by either the movement of the object or the sensor, considering the sensor's portability as either stationary or mobile, such as handheld scanners. For investigating deeper on the technology and the pattern used in SLS the interested reader can refer to %\citep{SALVI2010,Bell2016}; \citep[Chapter 6, p.~593-625]{Luhmann2020}
%[\citenum{Luhmann2020}, Chapter 6, p.~593--625;\citealp{SALVI2010,Bell2016}]
%[\citenum{Luhmann2020}, Chapter 6, p.~593--625; \citenum{SALVI2010}, \citenum{Bell2016}].
[\citenum{SALVI2010}, \citenum{Bell2016}] and [\citenum{Luhmann2020}, Chapter 6, pp.~593--625].

Short scanning time, high resolution and highly reduced dependency on material and texture are the key benefits of this approach. Hence, SLS is useful for surfaces with diffuse (lambertian) reflectors or lacking texture. Thanks to the fringe projection system, the device can provide high accuracy. Nevertheless, it is sensitive to the external surrounding light (illumination) and the sensor is usually utilized for mid-to-small objects and at a relatively close distance \cite{BUSWELL2022}. Additionally, SLS requires an internal warm-up, calibration, and temperature maintenance for optimal operation. For diving further into the source of error for SLS, the reader is referred to \cite{Rachakonda2019}.

Various handheld and stationary SLSs can be selected for QC in AMC. Handheld structured light systems such as Artec Eva \cite{artec3d} %(Fig XX b)
deliver 3D with submillimeter accuracy and spatial resolution. Although the accuracy and portability of these low-cost handheld SLSs is very good for most concrete 3DCP projects and allows the scanning of complex geometries. Even better results can be achieved using higher quality systems like AICON 3D Systems such as stereoscan neo \cite{Hexagon} %(Fig XX b)
, even though with much higher costs. This higher quality (accuracy and resolution) would be important in some applications like detecting fine defects on the concrete surfaces \cite{Backhaus2024}. In addition, the projectors offer back projection of the measurement results onto the scanned object.
%(see Fig. XXc)
This superimposition of the data is convenient for post-processing analysis and offers extended/mixed reality features. This enables repairs and adjustments directly on the object.

\subsection{Terrestrial laser scanner (TLS)}\label{TLS}
TLS has become a common technology in the acquisition of 3D point clouds in many engineering fields. Each measured point in the point cloud generally has 3D coordinates assigned with an intensity value and RGB values in recent scanners. The cartesian 3D coordinates are computed from the vertical and horizontal angles as well as the distance from the center of the scanner to the point. Additionally, the intensity value records the backscattered echo of the emitted signal \cite{LipkowskiMettenleiter2019}.  Most recent TLSs are integrated with built-in calibrated cameras with respect to their internal coordinate system with optical axes very close to the laser distance meter. Hence, the 3D point cloud can be back-projected to interpolate and assign a color value from the image in order to obtain a colorized point cloud, with minimum parallax effects.

The technology is based on either ToF or phase-shift principles \cite{Suchocki2020}. In some applications, triangulation-based scanners are used. The density, precision, and accuracy of TLS data could be affected by several factors and system configurations that are discussed comprehensively in \cite{Soudarissanane2016_Phd,Wujanz2016_Phd}.

\begin{figure}[h!]
    \centering
    \begin{tabular}{ccc}
        \includegraphics[width=0.3\textwidth]{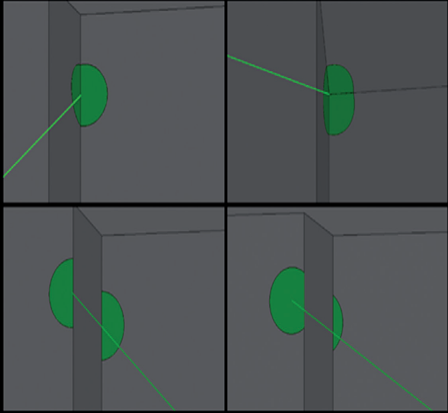}
        & 
        \includegraphics[width=.3\textwidth]{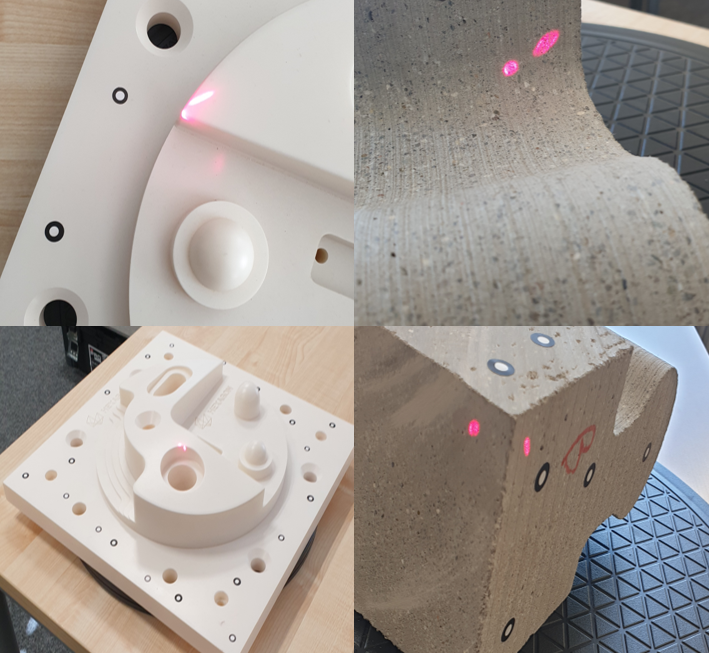}
        & 
        \includegraphics[width=.3\textwidth]{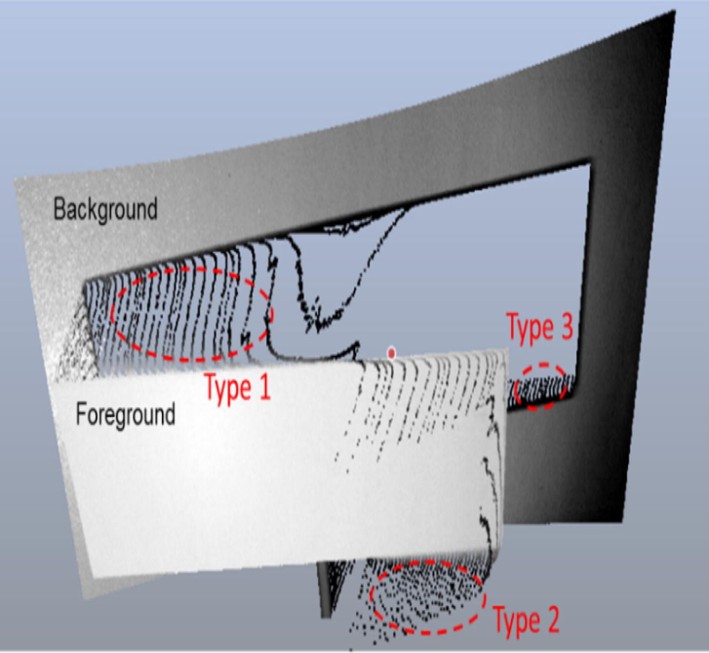} 
        \\
        a & b & c \\
    \end{tabular}
    \caption{Presentation of a variety of simulated and real laser beam’s footprints as well as the effect of mixed pixels in point cloud data. a) Simulated deformed beam footprint on edges, image credit: \cite{Klapa2017}. b) Real deformed beam footprint on corners and edges for different objects. c) Mixed pixels result in three different types: Type 1 is between the background and foreground surfaces. While Type 2 points are in front of the foreground surface. Finally, Type 3 points are behind the background surface, image credit: \cite{WANG2016_mixedpixels}.}
    \label{fig:Laser beam deformation and Mixed Pixels}
\end{figure}

Sensor warm-up is necessary for a stable and reliable measurement \cite{Janssen2021}. The authors indicate that there is drift in the measurements during both the warm-up phase and long-term usage. Thus, the author proposes that the scanner be warmed up for a minimum of one hour prior to use, given that during the aforementioned phase, a drift of up to 0.2 mm is observed for spherical components (vertical and horizontal angles, and range).
Although laser scanning offers high accuracy in 3D measurements and low dependency on lighting conditions, it faces limitations in material reflectivity and a decrease in point density with increased sensor-to-object distance. Furthermore, the pseudo-random spatial sampling of TLS hinders the direct capture of object edges. Also, edges and corners generate errors due to multi-echo effects, where the laser beam's footprint divides over the edge, recording mixed signals from adjacent surfaces, leading to \textit{'mixed-pixels'} that affect distance measurements (See Fig.~\ref{fig:Laser beam deformation and Mixed Pixels}).

In addition, this phenomenon is caused by the beam's footprint changing from a circle to an ellipse, which is a function of the angle under which the beam hits the surface (incident angle, IA). This shape deformation would shift the center of the footprint away from the line of sight used for angle measurements. As a result, the localization of the point is error prone \cite{boehler2003,Klapa2017,ZAMECNIKOVA2018,Chaudhry2021}.
The phenomena are explained further by \cite{ZAMECNIKOVA2018} to illustrate the systematic deviation of measurements phenomena caused by IA into two sources of errors: (i) the geometry of laser footprint geometry is deformed which results in different distances, depending on the size of the beam diameter at the target and the orientation of the surface, (ii) under higher IA the received strength of the signal is reduced.

Free-form printing in modern concrete fabrication methods like AMC and the possibility to produce concrete objects with complex geometry, data capturing of wet concrete e.g. in stage-wise QC (cf. sec.~\ref{sec4: Quality Control}), time limitations within the printing process, and the existence of different material in one object, necessitate special considerations when employing TLS in AMC.

\section{Data Capture}\label{sec3: Data Capturing}
Like other engineering problems, optimizing the time and cost while reaching the required quality (precision, accuracy, and LOD) is important in spatial data capturing for construction purposes. Stationary and temporal clutters, moving objects, low texture and lighting conditions, safety issues, weather conditions, and time limitations are some parameters, among others, that make data capturing at construction sites challenging \cite{ZHANG2016,Vincke2019,Maboudi2023}. In this section we will discuss the effects of objects’ properties and environmental conditions on data capturing. Afterwards, we will cover data capture planning and registration of the acquired data.

\subsection{Objects’ properties and  environmental conditions}\label{subsec3.1:Object_and_Env}

Several factors may affect the quality of the information acquired by optical sensors. This begins with the wavelength of the signal that we aim to omit/receive with our sensor. In the case of terrestrial photogrammetry, for example, this is typically the visible light range. The entire measurement processes - from the preparation of the object of interest, to image acquisition (such as controlling the exposure), and finally to image analysis - all play a role in determining image quality \citep[Chapter 3]{Luhmann2020}.

\begin{figure}[h!]
    \centering 
    \begin{tabular}{cc}
        \includegraphics[width=0.40\textwidth]{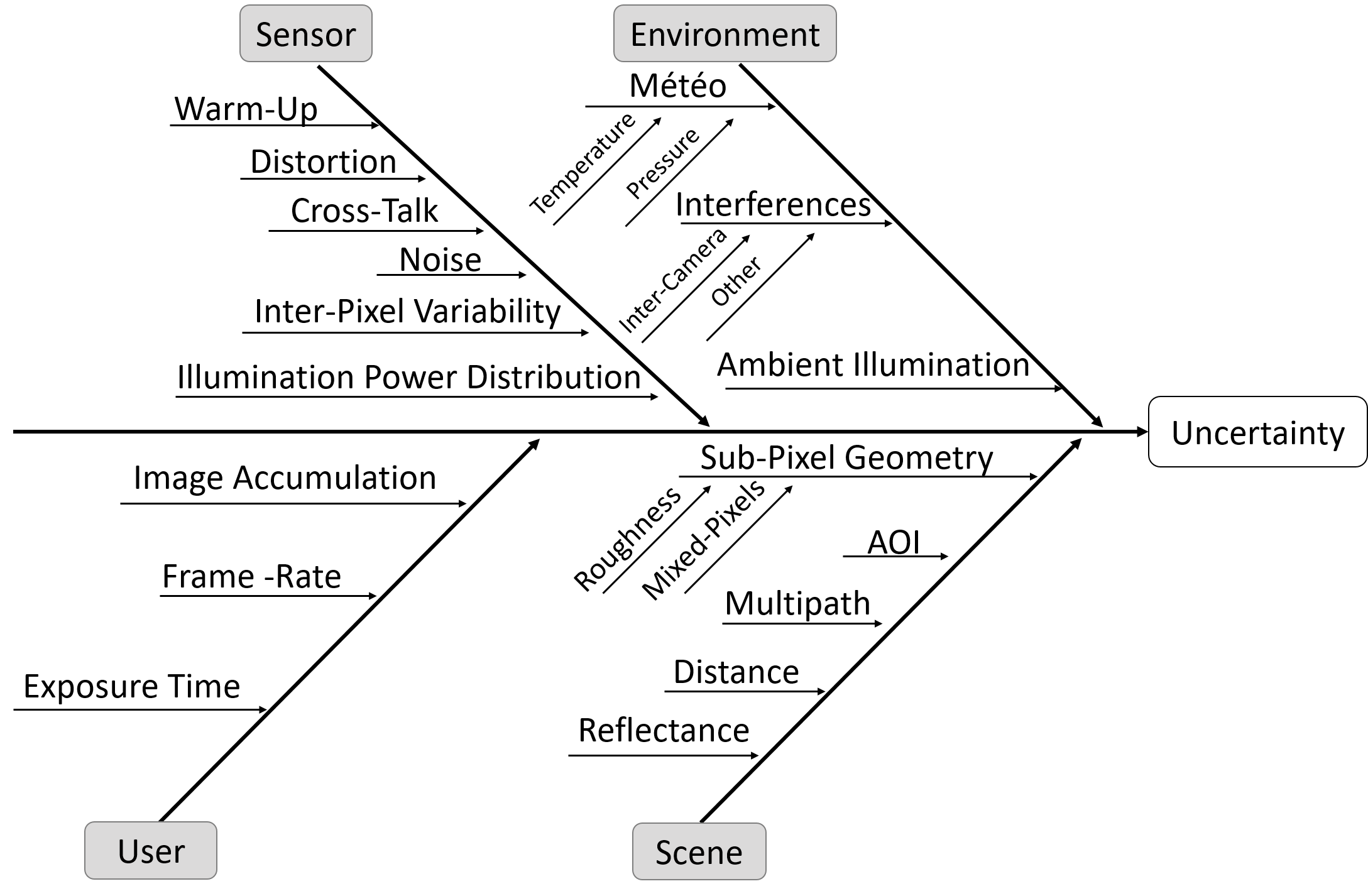}
        & 
        \includegraphics[width=.54\textwidth]{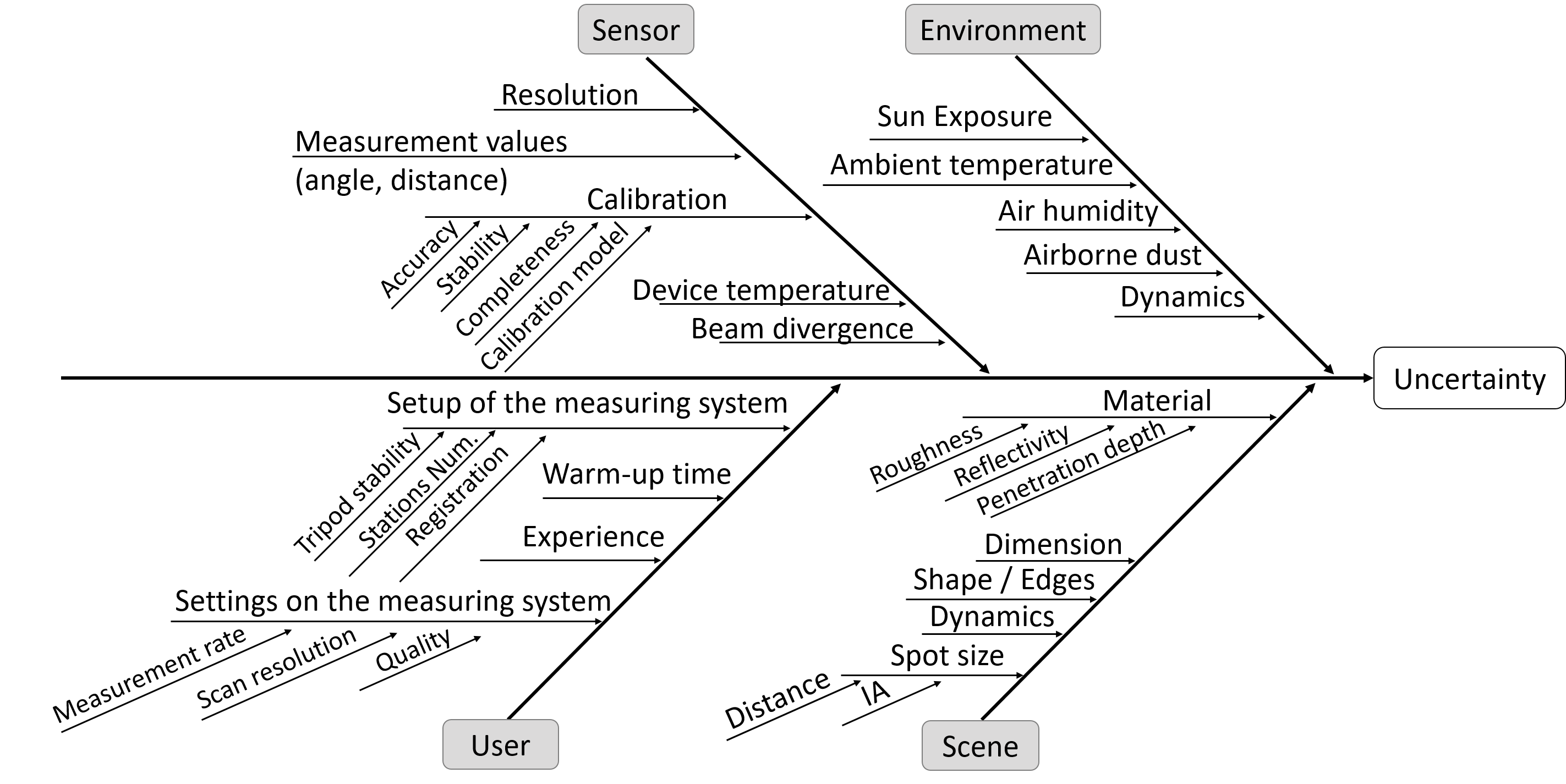}
        \\
        a & b \\
    \end{tabular}
    \caption{Ishikawa diagrams for the uncertainty of measurement: a) Diagram for ToF camera, Credit: \cite{Frangez2022}. b) Diagram for TLS, image adopted from \cite{LipkowskiMettenleiter2019, Hartmann2023}.}
    \label{fig:Ishikawa}
\end{figure}

Every sensor has its own source of uncertainty, which can be traced using an Ishikawa diagram. An overview of the sources of error with respect to ToF cameras is provided in \cite{Frangez2022}, as shown in Fig. \ref{fig:Ishikawa}a. For TLS, there are four main groups of error sources: instrumental imperfections, environment, scan geometry - including incident angle and range to the surface, and object properties \cite{SOUDARISSANANE2011, Soudarissanane2016_Phd}. An Ishikawa schematic for TLS sensor is provided in \cite{LipkowskiMettenleiter2019, Hartmann2023}, as shown in Fig. \ref{fig:Ishikawa}b.

To ensure that this paper aligns with the scope of AMC, relevant aspects that are specifically important for AMC will be addressed. For further and more detailed information, necessary references will be provided. 
\begin{comment}
The discussion will cover the following parameters: Environmental conditions, and Object influences (Shape, Size, Range, IA).
\end{comment}

\subsubsection{Environment}\label{subsubsec3.1.1:Env.}

The environmental condition in AMC is an important factor that must be considered in order to achieve high-quality data. The printing material used for shotcrete, for example, is mainly based on cement and water. Concrete material spreads all over the site. Moreover, any material that spreads can cause direct damage to the sensor by scratching its lens. Thus, an investigation into lens shield protection is essential for maintaining the sensor's health while ensuring that the signal does not deteriorate. Nevertheless, this might be more important for online data capturing than for post-printing inspection. Additionally, a special concern needs to be considered related to the temperature and humidity effects on the data capturing quality \cite{Zogg2008,Maboudi2020,Frangez2022_feedback}.

For instance, the operating temperature of the Z+F IMAGER\textsuperscript{\textregistered} 5016 laser scanner,  varies from -10° to + 45°C and also it is recommended to work in non-condensed humidity \cite{ZF5016}. In general, the data capturing step could proceed after some time from the printing splashing in case of no aeration. The data might be acquired using freshly produced concrete surfaces, therefore, understanding the implications of fresh concrete on the quality of the data acquisition is necessary. The effects of measuring wet concrete surfaces were pointed out in literature for TLS applications and images \cite{Suchocki2018,Garrido2019}. The effect of the water layer on the rough concrete surface (saturation) affects the light signal by causing internal reflections and absorption of pulses within the water-air layer, resulting in a darker surface than a dry one \cite{Suchocki2018}. This is true for both images and TLS. Furthermore, the TLS signal is affected in two ways, namely (i) a reduction in the returned pulse power, and (ii) an increase in the specular reflection of the laser beam and a reduction in the diffuse reflection.

Shifting focus from the object to the surrounding environment, specifically the construction site for the scope of AMC, \cite{Vincke2019} investigate the challenges associated with photogrammetric data capture in construction. However, these challenges can primarily be attributed to various sensors. The paper addresses issues such as occlusion and moving objects, reflections and poor surface textures, narrow spaces, and inadequate illumination. The author explains the difficulties posed by equipment and clutter on construction sites, which can include molds, scaffolding, personnel, and machinery. These challenges impact the effectiveness of registration between two datasets, particularly in cases of occlusion or when objects are moved.

\subsubsection{Object (Shape \& Size)}\label{sec3.1.2:Object (Shape and Size)}
\begin{figure}[h!]
    \centering
    \begin{tabular}{ccc}
        \includegraphics[width=0.35\textwidth, height=0.1\textheight]{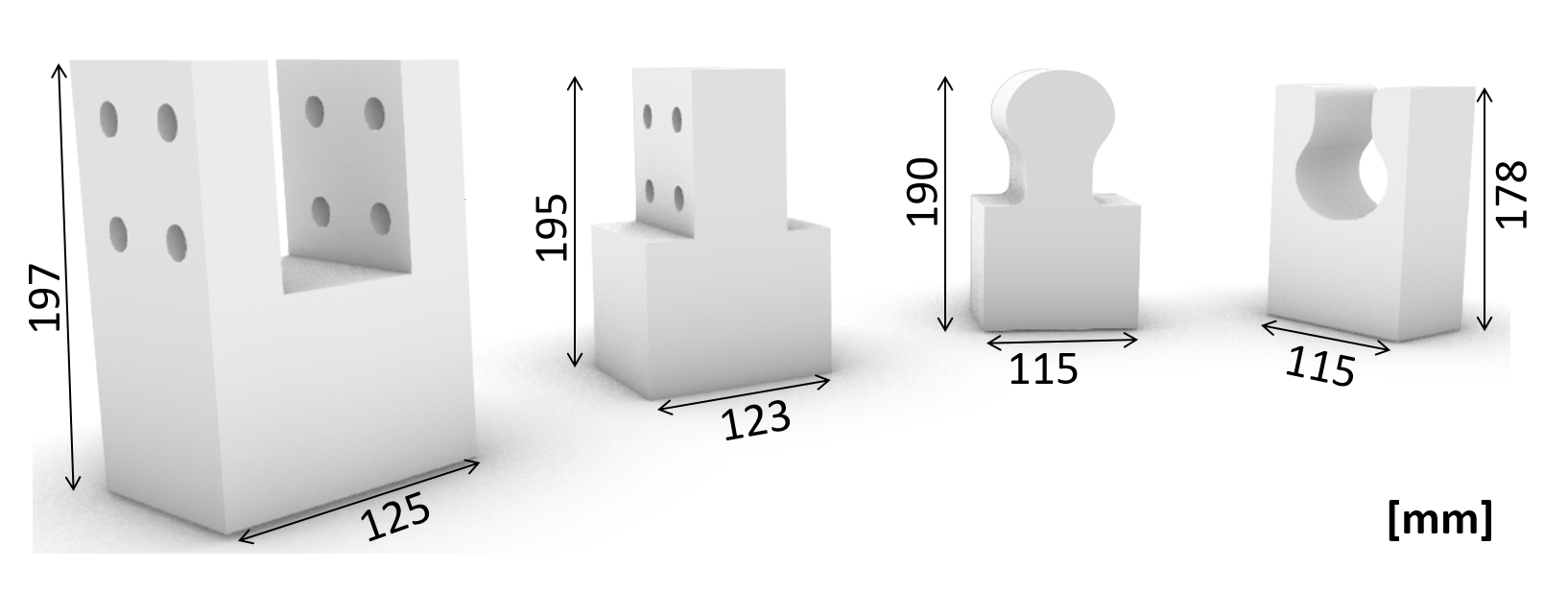}
        & 
        \includegraphics[width=.25\textwidth, height=0.15\textheight]{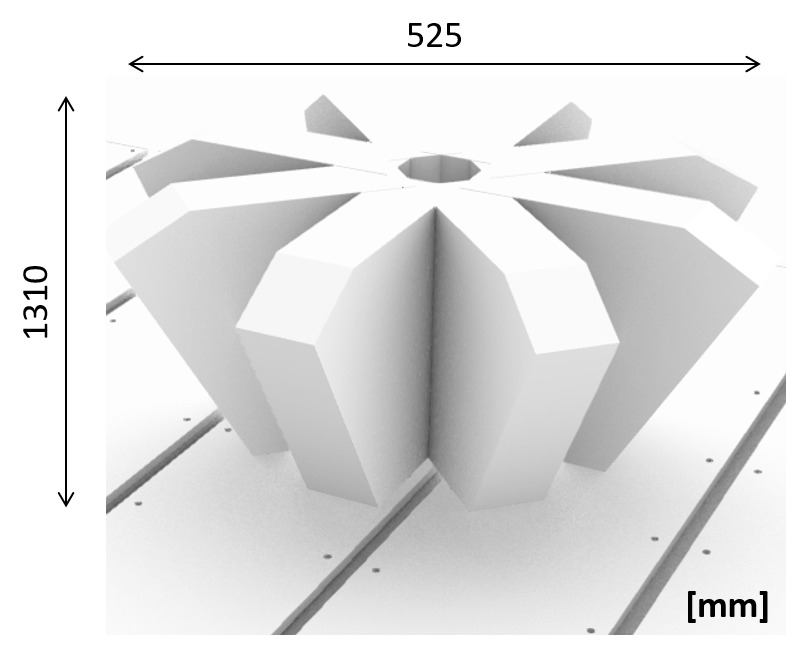}
        & 
        \includegraphics[width=.3\textwidth, height=0.2\textheight]{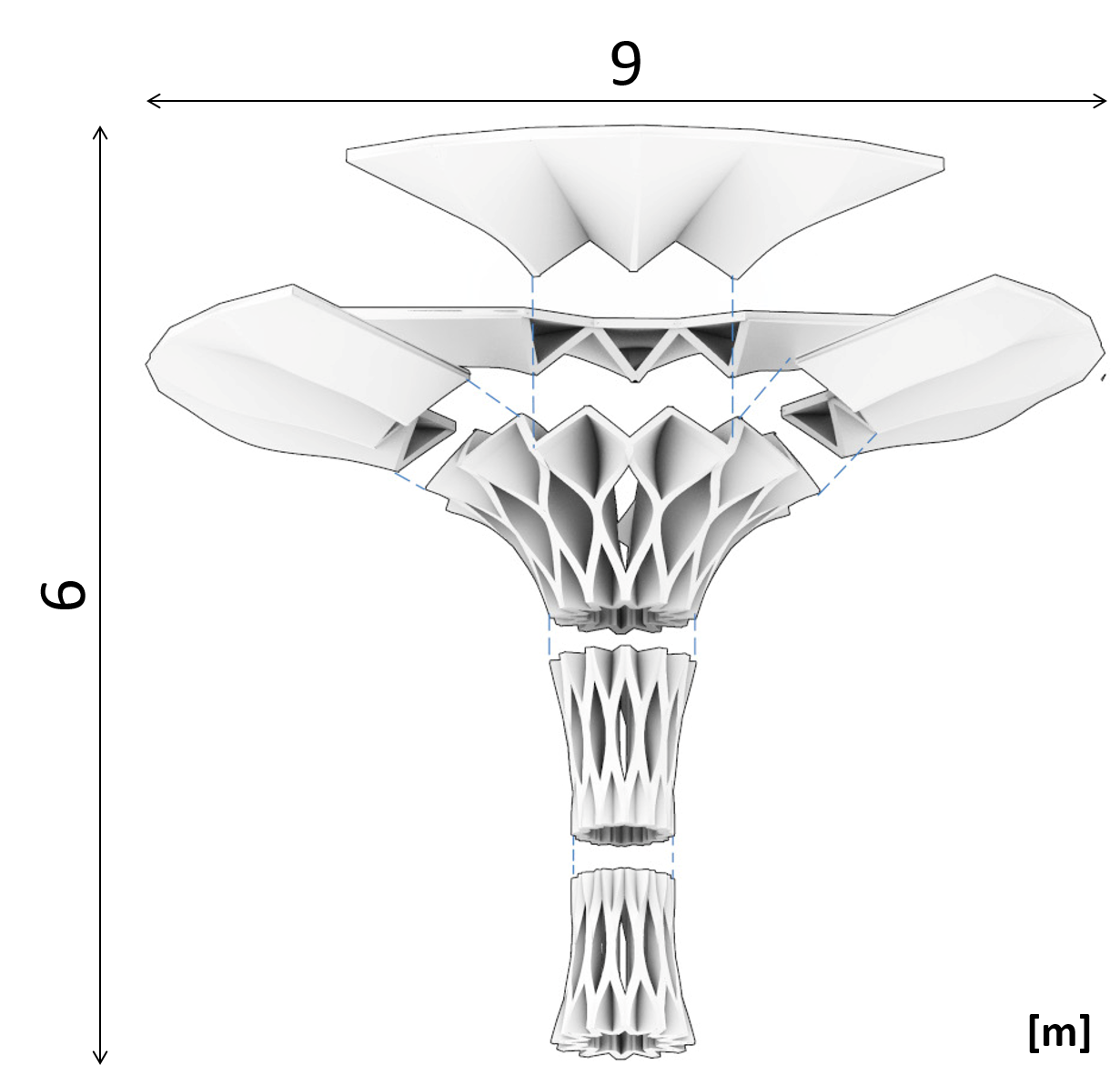}
        \\
        a & b & c\\
    \end{tabular}
    \caption{Designed models of the objects with different shapes and sizes. a) All of the objects are approx. in a size of 100 mm $\times$ 200 mm in width and height, respectively, b) A Star-shaped object with challenging angles and grooves, c) A conceptual model of a segmented calyx-like column, image credit: \cite{Kloft2019}.
}
    \label{fig:Objects}
\end{figure}

The influence of objects properties like shape and size on data capturing has been thoroughly investigated by several researchers, including \cite{Kersten2005, boehler2003,Klapa2017,Maboudi2020}. Since optical sensor measurements are limited to the line-of-sight observations, shape complexities which are very common in AMC (see Fig.~\ref{fig:Objects}) affect the data capturing. Hence, tackling this problem requires a special concern in order to generate complete and high-quality data \cite{Biswas2015}.

Intrusions, extrusions, and different sizes of the objects in Fig.~\ref{fig:Objects} poses several challenges for reality capturing. For example, TLS stations or the position of depth-camera should be cautiously planned so that all parts of the objects are visible with a proper incidence angle and range to meet the quality requirements \cite{SOUDARISSANANE2011}. In the case of image capturing, SfM and MVS heuristics (cf. sec.~\ref{subsec_3.2.2: Data capture planning for photogrammetry}) should also be respected. These necessitate proper data capture planning which is the topic of the next section (cf. sec.~\ref{subsec_3.2: Data capture planning}).

In correlation to object geometry, range, and IA, two further factors must be considered when assessing the quality of captured data. IA, the angle between the laser beam / beam of light and the local surface normal, should be minimized \cite{SOUDARISSANANE2011,Turley2014}. This hold true for all discussed sensors namely: cameras, SLS, and TLS. Additionally, IA exhibits a correlation with the surface roughness, affecting how the signal is reflected. Rough surfaces increase the area of the laser beam footprint. Additionally, surface roughness can lead to multiple reflections of emitted light from the sensor, thereby increasing the intensity value of the back scattered received signal \cite{Suchocki2018}.

\subsection{Data capture planning}\label{subsec_3.2: Data capture planning}
Effective data capture planning is crucial for ensuring data quality and efficiency. Main aspects of planning the data capturing procedure for TLS and photogrammetry will be discussed here. Similar approaches could be generalized for depth capturing using other conceptually similar sensors like SLS and depth cameras. A basic idea about efficient data capturing could be formulated as a set-cover problem or an art gallery problem \cite{Vincke2019}, where the main objective is to cover the area of interest with minimum viewpoints. Additionally, LOA, LOD, and LOC metrics should be considered in data capture planning \cite{ARYAN2021}. Thereby, a thorough and suitable plan is required to ensure adequate density and coverage for all the required details on the object \cite{Biswas2015}. The following sections will discuss the prerequisites for both TLS and photogrammetry, as well as the current state of digital construction.

\subsubsection{Data capture planning for TLS
}\label{subsec_3.2.1: Data capture planning for TLS}
Several different aspects have to be considered when aiming to find the optimal positions of a TLS. Generally, this problem which is also known as plan for scanning (P4S) optimizes the following aspects:
\begin{enumerate}
  \item[a)] Minimum number of standpoints; to reduce the data capture and later processing time.
  \item[b)] Visibility analysis to cover the object of interest or the scene, effectively.
  \item[c)] The quality of the point cloud per standpoint, which is heavily related to the measurement configuration.
\begin{comment}
    The incidence angle of the beam with regard to the measured object and the distance from the scanner have a big impact on the measurement quality.\mg{quality: in which respect? In the end it is first of all point density, but this is not necessarily a quality indicator, depends on the object properties. The point accuracy is also distance dependent, but in certain ranges noncritical , KM:The quality in general about the metrics LOD, LOA and LOC}  
  Thus, these scanner-related constraints have to be considered, given the manufacturer's restrictions and recommendations.
\end{comment}

\end{enumerate}
A set of parameters must be defined for optimization, while another must be predefined as a constraint for the optimization algorithm to search for the optimal combination of parameters for the existing constraint and object, with the goal of reaching the defined objective. Some parameters like the height of the scanner and angular resolution are usually excluded from the optimization process \cite{Noichl2024}. However, some researchers like \cite{ZHANG2016}, tuned the angular resolution as a parameter in the optimization, which enhanced the effectiveness of the data at the cost of increasing the complexity of the optimization. Nevertheless, as emphasized in \cite{Wujanz_2016},  the necessity of conducting a simulation prior to the actual optimization process emphasizes the importance of considering the specifications of the TLS and the object's shape from predefined viewpoints.

Furthermore, prior to the implementation of an optimization strategy, some researchers generate a set of candidates for evaluation, though the approaches utilized for this generation vary. Researchers have addressed the issue of candidate generation through the implementation of a variety of methods. One straightforward approach is to generate a grid search for candidates with fixed spaces as described by \cite{Dehbi2021}. Alternatively, candidates are generated by a possession disk sampling, which ensures that points are evenly distributed across the surface, regardless of the object's shape or complexity, as outlined in \cite{Noichl2024}.

Most researchers use the greedy algorithm (GA) as their optimization function combined with next best view (NBV) strategy to minimize the time for scanning \cite{Chen2018, ARYAN2021}. GA is an iterative process that employs a heuristic technique to achieve a local optimal gain at a single stage \cite{Chvatal1979}. Two different strategies, namely: the optimal heuristic strategy (two-phase optimization) and the NBV strategy combined with weighted GA strategies, are compared in \cite{Li2022}. The results demonstrated that NBV exhibited significantly higher efficiency than two-phase optimization. However, the two algorithms exhibited disparate behavior with regard to the coverage of the construction site. Specifically, the two-phase optimization method demonstrated superior performance in covering the central area of the site relative to the weighted GA.

For small and conventional projects, path planning for TLS is trivial and the operator can decide on that, some researchers addressed this problem, mostly based on Traveling Sales Problem (TSP). Greedy algorithm with TSP are applied for P4S after generating the selected candidate of the viewpoint-planning (VPP) for path planning generation in \cite{Noichl_and_Bormann2024}.

\subsubsection{Data capture planning for photogrammetry
}\label{subsec_3.2.2: Data capture planning for photogrammetry}
Generally, in camera based data capture solutions, the Field-of-View (FoV) is notably different in comparison to TLS, which usually provides a panoramic view of the scene. It is constrained by the specifications of the camera, namely focal length and sensor size. This limitation also applies to both SLS and depth cameras. Nevertheless, even with the utilization of 200° (viewing angle) Fish-eye lens, a difference in field of view exists between TLS and camera systems.

The following main constraints should be considered for VPP are (cf. sec.~\ref{Photogrammetry}):
\begin{enumerate}
  \item[1.] GSD (Ground Sampling Distance): This parameter which reflects the object space resolution, affects the distance between the camera positions to the object, given a pre-defined focal length.
  \item[2.] Sufficient overlap between the images to capture each point on the surface of interest from at least two images.
  \item[3.] Observation angle, i.e. the angle between the direction of the optical axis and the observed surface, and intersection angle, which is the parallactic angle as the difference between adjacent cameras. As a rule of thumb, the observation angle should be bigger than 20° for complicated objects. Moreover, the suggested value for the intersection angle is about 90°-100° to provide a reliable depth estimation \cite[p.~650]{Luhmann2020}.
\end{enumerate}
\begin{comment}
Terrestrial photogrammetry is a well-studied problem that has been tackled and investigated by researchers 
[\citenum{Fraser1984},\citenum{Forstner2016}, Chapters 11--15].
\end{comment}
Terrestrial photogrammetry is a well-studied sub-domain of photogrammetry that has been tackled and investigated since the 80s [\citenum{Fraser1984}] and [\citenum{Forstner2016}, Chapters 11--15]. Optimizing an already established network for finding the best view points to enhance the accuracy and effectiveness is proposed in \cite{Saadatseresht_and_Varshosaz2007}, by utilizing an artificial neural network. Similarly in \cite{Alsadik2014}, the authors proposed methods to optimize camera networks, ensuring coverage and accuracy while reducing the number of cameras needed. A wide range of papers are examined in \cite{Maboudi2023}, exploring novel approaches and techniques for VPP on agile platforms, such as UAVs. The next section is focused on the employment of mobile robots as a sensor platform for digital construction planning as well as the discussion of construction characteristics.

\subsubsection{Data capture planning in digital construction}\label{subsec_3.2.3: Data capture planning in digital construction}
The search for best view planning methods are divided into two categories a) Model-free based, and b) Model-based \cite{Maboudi2023}. A close examination of the characteristics of digital construction reveals that the exploration followed by exploitation (ETE) strategy is more aligned with the requirements of AMC. Since the BIM or as-designed 3D model of the objects are already available in digital construction, this gives an opportunity to do the network design even before visiting the site and printing the object.

\begin{figure}[h]
    \centering
    \begin{tabular}{cc}
        \includegraphics[width=0.58\textwidth]{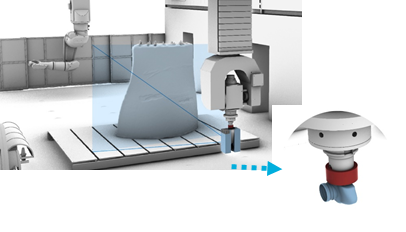}
        & 
        \includegraphics[width=.33\textwidth]{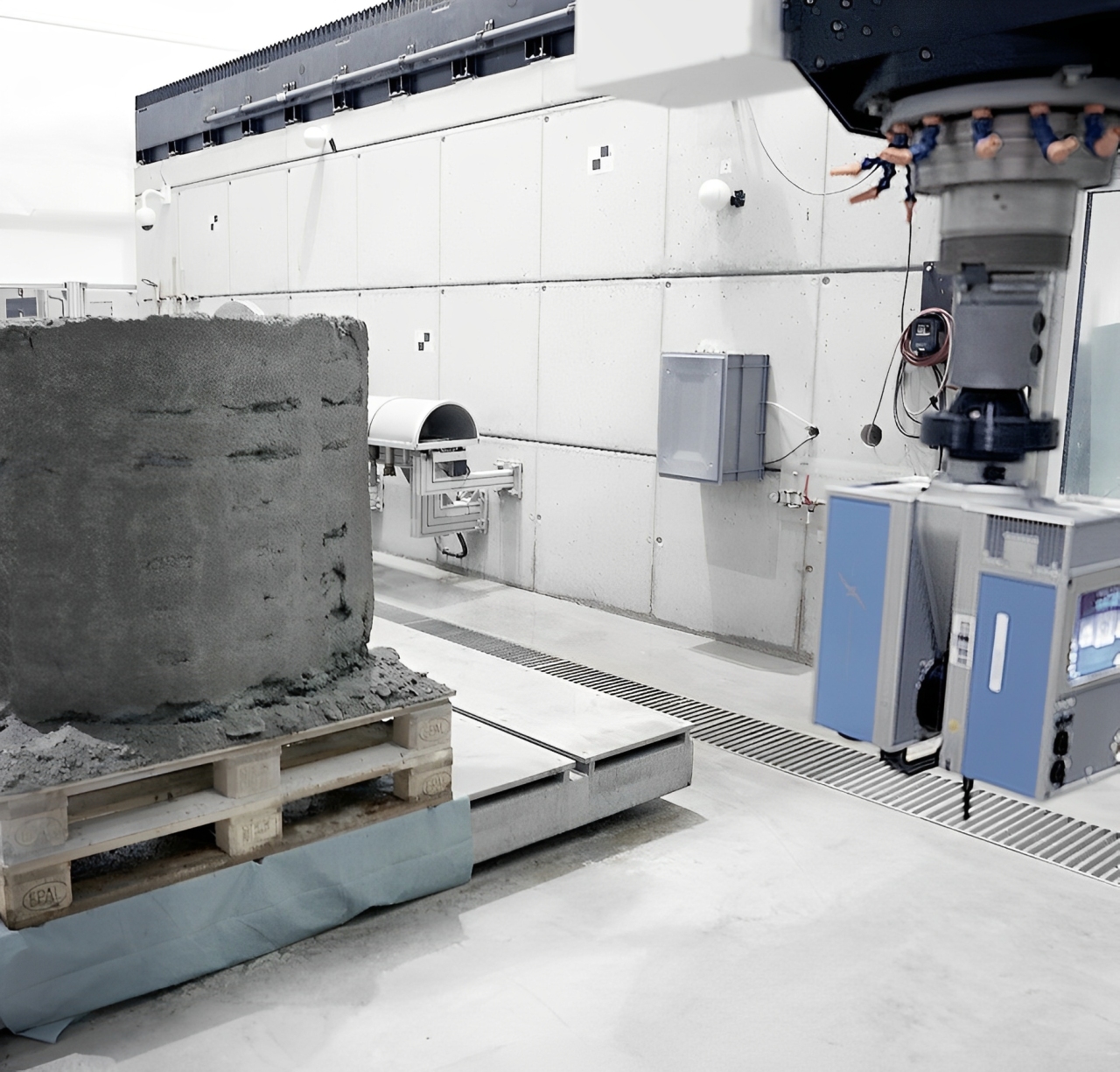}
        \\
        a & b \\
    \end{tabular}
    \caption{ Sensors mounting on robot for AMC. a) concept, b) realization. Images credit: \cite{Mawas2022_Automatic-Inspection, AMC-TRR277}.
}
    \label{fig:Mounted Sensors}
\end{figure}

Model-based approaches can be utilized when the input model represents the scene \cite{ARYAN2021,Maboudi2023}. However, a construction site is a very dynamic scene with many permanent and temporal clutters \cite{Vincke2019}. Hence, mismatches between the model and reality may produce errors. Additionally, after computing the optimal sensor pose, approaching those may not be possible in conventional data capturing solutions. Nevertheless, in digital fabrication one can cope with this challenge with a sensor mounted on a robot arm (see Fig.~\ref{fig:Mounted Sensors}). Mounting the data capturing sensor on the fabrication robot arm offers higher degrees of freedom. For example, the flexible sensor height can be included as a parameter in the optimization process. In the case of on-site or in-situ fabrication, UAVs and mobile robots can be employed as  platforms carrying the sensors to the computed pose \cite{Oelsch2021, Vega-Torres2024}.  Furthermore, other factors like shadow and illumination must be considered when working with passive sensors, as well as challenges related to reflectivity or low-texture surfaces \cite{Maboudi2023}.

Another aspect that could be considered during VPP is the optimization of the target's position that will be considered to connect the sensors and thereby the acquired data. This idea was adopted in \cite{Jia_and_Lichti_2019}, combined with the VPP process. Moreover, the target-based data capture planning enables the direct registration of point clouds. This method significantly reduces the registration errors which are discussed in sec.~\ref{subsec3.3: Registration}, which are typically considered as the largest factor affecting the accuracy of the data.

\subsection{Registration}\label{subsec3.3: Registration}
After data capturing, the process of overlaying the captured data sets and also the as-designed model \textendash each with different coordinate systems\textendash\ into a common coordinate system represents the essence of data registration \cite{Tondewad_and_Dale_2020}. 
\begin{comment}
Registration is an old problem tackled by engineers which can be traced back to the
early 70s in remote sensing for Multispectral 2D data \cite{Anuta1969, Anuta1970}
\end{comment}
The task of registration is tackled in different disciplines namely photogrammetry and remote sensing \cite{Tondewad_and_Dale_2020, Cheng2018_review}, computer vision \cite{Yang_TEASER_2021, Lim2024_KISS-ICP}, mobile robotics \cite{Pomerleau2015, Vizzo2023_KISS-ICP}. In AMC, also, due to the availability of different sensors, efficient and accurate registration of data sets is crucial for any further analysis and inspections. A compromise between the efficiency and accuracy of co-registration is expected and tolerated as long as the fabrication tolerances are not violated.

A variety of data capturing sensors are integrated with both an Inertial Measurement Unit (IMU) and a Global Navigation Satellite System (GNSS) receiver, enabling the estimation of the sensor's pose and facilitating direct sensor registration. However, in GNSS-denied environments, e.g. off-site AMC, the sensor can be mounted on a robot to obtain its pose directly from the end effector of the robot (see Fig.~\ref{fig:Mounted Sensors}). On the other hand, indirect registration using signalized targets is a prevalent method for dataset alignment, facilitating transformations like 3D similarity for registration tasks \cite{Cheng2018_review}. For laser scanning point clouds, target design and detection algorithm for estimating the position \cite{Janssen2019} or consistency of fixed-targets orientation in different scans \cite{Janssen2021} can affect the results.

There are different iterative approaches for solving the registration problem without signalized targets. A general approach could be a coarse registration followed by a fine registration \cite{PLADE2020}. A very well-known fine registration approach is the iterative closest point (ICP) algorithm which was first introduced in \cite{Besl_and_McKay1992}. The pipeline of ICP consists of five crucial steps \cite{Rusinkiewicz_and_Levoy_2021}: (i) Selection of correspondences, (ii) Matching of correspondences, (iii) Weighting of correspondences, (iv) Rejecting wrong pairs, and (v) Error metric (cost function) minimization. 

It is worth noting that the original ICP algorithm uses the entire point cloud. When there is little pose difference between the two datasets or when the coarse alignment is available, ICP can converge to the global optimum. However, ICP is known to be vulnerable to outliers, local minima in case of self-similarity of the point clouds or if the coarse registration is not sufficiently close to the optimum\cite{Mawas2022_Direct_Coregistration, Lim2024_KISS-ICP}. 
Also, local failures in the fabrication process will be averaged using ICP \cite{Mawas2022_Direct_Coregistration}, which renders a comparison of captured data to the model uncertain. Furthermore, ICP is inefficient for some applications when time is crucial. Therefore, different variants of ICP are introduced in the literature to overcome or mitigate the shortcomings of naive ICP \cite{Yang2016_GO-ICP,Zhou2016_FGR-ICP,Zhu_2019,Yang_TEASER_2021,Lim2024_KISS-ICP}. In general, the surface geometry in the overlapping area must resemble at least three primary normal directions that are not parallel in order to allow proper registration and convergence to the correct optimum \cite{Mawas2022_Direct_Coregistration, Lim2024_KISS-ICP}
\begin{comment}
\mg{Also the issue with error-averaging is very important in our context: an inherent assumption of ICP is that objects represented in both clouds are  idententical. If this is not given, subsequential QC can be biased (see own paper). It seems this registration section refers to the registration of captured data to model. This is fine, but then the named issue is a very important restriction!}.     
\end{comment}

Plane-base registration has proven to be a common approach to solve the registration task, as well as to enhance the robustness of registration by adding several plane correspondences from the surrounding scene\cite{BOSCHE2012,Scantra2018, Zhao_PlaneBased_Registration2022}. While the details of various registration algorithms are out of scope of this paper, the interested reader can refer to \cite{Salvi2007, Cheng2018_review, Dong2020, Zhang2020, Huang2021, Chen2024, Zhao_and_Fan_2023,MAISELI2017}.

\section{Quality Control}\label{sec4: Quality Control}
In the manufacturing sector, geometric quality control is of paramount importance for guaranteeing that products comply with design standards, thereby facilitating automation and enhancing efficiency.
For non-flat printing bases (substrates), the very first step of QC starts before fabrication to capture the existing printing bed \cite{Nicholas2020}. In the case of standard substrates, pre-quality control, and path planning simulations are required. Before printing, using the simulation or test prints, some geometrical aspects could be analyzed \cite{Zhan2021, Breseghello2021,Lao2020}. A depth camera aligned on the extruder is utilized in \cite{Naboni2022} to collect geometric data of the existing substrate to update the path planning and after printing comparing as-built point clouds to the digital model. In SC3DP, depth cameras are employed by Frangez et al. \cite {Frangez2021} to detect the reinforced mesh in order to finally facilitate path planning. A mobile robot is equipped with a depth camera and a 2D laser profiler for infrastructure repair in \cite{Dielemans2024}. The robot, which is initially operated by a human, scans the target area to create detailed surface reconstructions, enabling precise path planning for repairs. These strategies highlight the role of pre-processing for adaptive design. In contrast, unknown substrates depend on real-time data capture and adaptive design. Incorporating sensor feedback facilitates enhanced environmental awareness, thereby aiding the adaptation of tool paths \cite{Wolfs2018, Naboni2022, Capunaman2023}.

\begin{figure}[h!]
    \centering
        \includegraphics[width=.7\textwidth]{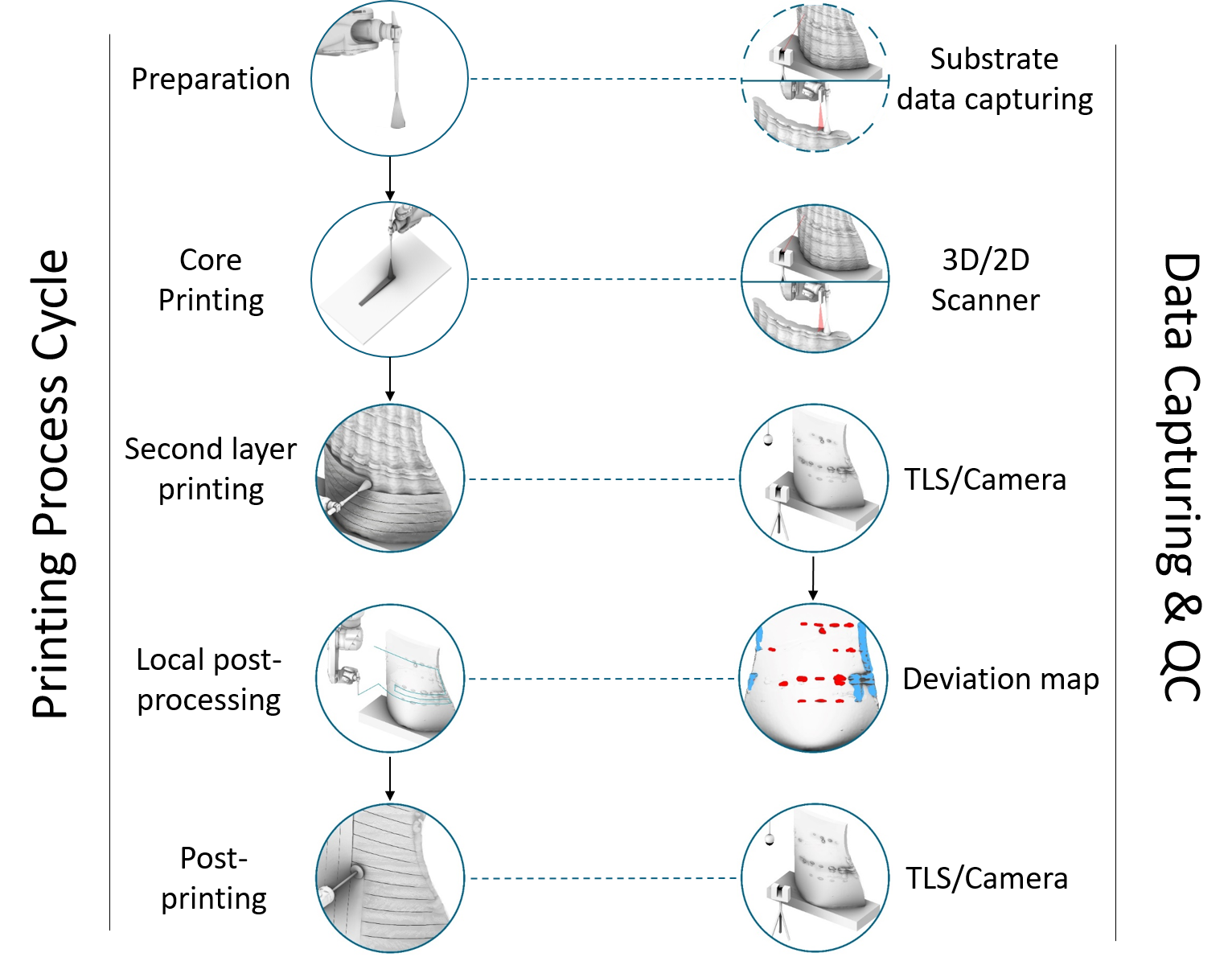}
    \caption{ Printing object workflow integrated with QC. Partially adopted from \cite{Maboudi2020}.} \label{fig: QC_workflow}.
\end{figure}

In AMC, hybrid fabrication methods combining additive and subtractive processes are of particular importance, especially when changes in material properties over time, such as shrinkage, impact the results \cite{Mendricky2023}. Hybrid fabrication methods have been demonstrated to mitigate these effects \cite{Hack2022}. The tool path planning process, which is informed by quality control analyses, guides each fabrication stage. This is done to ensure conformity between the objects that are printed and the digital models that are used as a reference (see Fig.~\ref{fig: QC_workflow}).

QC can be applied during different stages through the AM cycle. Several researchers,\cite{Buswell2020, Mawas2022_Automatic-Inspection, Slepicka2024, Farrokhsiar2024, WOLFS2024} classified QC based on its application throughout the variety of the fabrication stages. Hence, we categorize QC to stage-wise QC of four categories, namely into: (i) During Printing (Online), (ii) Layer-wise, (iv) Pre-assembly, and (v) Assembly, which are investigated in the following subsections.

\subsection{During printing (Online)}\label{subsec 4.1: Online}
In recent advancements in 3D concrete printing, various studies have explored methods for printing monitoring and control to enhance print quality. Continuous adjustment of the nozzle height relative to the surface using 1D Time of Flight (ToF), is introduced in \cite{Wolfs2018}. The sensor is placed in front of the nozzle to enable real-time measurement, which improved print accuracy. Online measurement approaches for SC3DP by comparing a ToF depth camera with 2D profile laser triangulation are investigated in \cite{Lindemann2019}. This area is further advanced by presenting a control unit for inline process evaluation and control of SC3DP in \cite{Lachmayer2022}. The authors utilized a forward-looking 2D laser profiler for assessing the nozzle-to-filament distance and layer width to adjust printing parameters like speed and spray distance. Conversely, a backwards-looking 2D profiler for instant measurement of filament width after printing is employed in \cite{WOLFS2024}. Additionally, for SC3DP,
four ToF depth cameras, to evaluate the vertically-printed surface based on local geometric features and intensity values, are employed in \cite{Frangez2022_feedback}. The system also assessed material thickness by comparing as-built and as-designed states or different stages of as-built progress.

\begin{figure}[h!]
    \centering
    \begin{tabular}{cc}
        \includegraphics[width=0.7\textwidth, height=0.13\textheight]{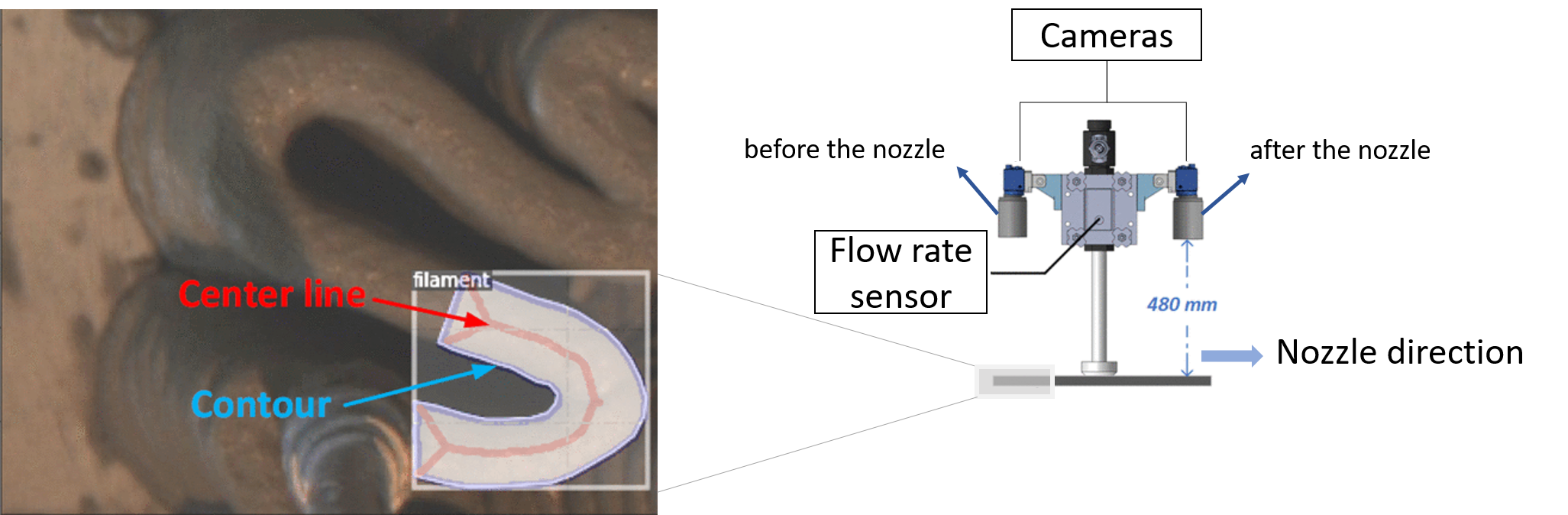}
        & 
        \includegraphics[width=.2\textwidth, height=0.13\textheight]{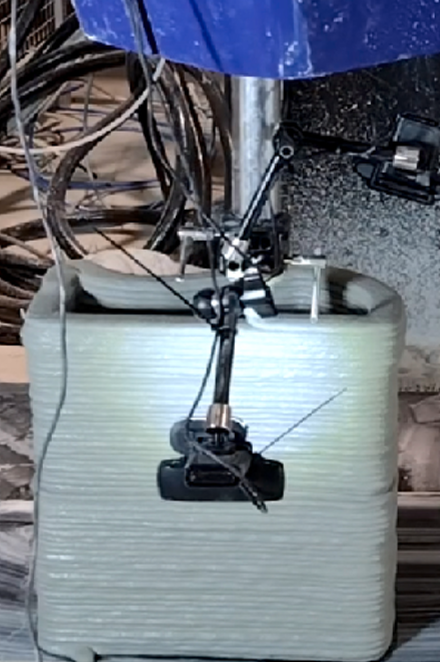}
        \\
        a & b \\
    \end{tabular}
    \caption{ Various camera constellations in relation to their viewing directions. a) Top-view direction, image modified from: \cite{Yang2022, Yang2024}, b) side view, image credit: \cite{Rill-Garcia2022}.
}
    \label{fig:Sensors Positions}
\end{figure}

Techniques for real-time extrusion monitoring are examined in \cite{Kazemian2021}. The study indicates that vision-based approaches are very promising for the inspection of filaments’ geometry. Mask R-CNN, for instance segmentation, is developed in \cite{Yang2022, Yang2024} for automatically detecting filament width deviations from camera images before and after the nozzle (see Fig. \ref{fig:Sensors Positions}a). a camera perpendicular to filaments (see Fig. \ref{fig:Sensors Positions}b) for height monitoring, using U-Net for segmentation is employed in \cite{Rill-Garcia2022}.

\begin{table}[h]
\caption{Overview of sensor integration in AMC inspection processes during the printing stage. (/) indicates that the information is not available for that particular entry.}\label{Table1_online}
\setlength{\tabcolsep}{3pt} % Adjust column separation
\renewcommand{\arraystretch}{1} % Adjust row separation
\small % Reduce font size for the table
\begin{tabular}{lllllll}
\hline
Paper         & AM        & Sensor                                                                                           & Position                                                                 & Direction & Inspection                                                                              & Control                                                                                                                                                    \\ \hline
\cite{Lindemann2019}      & SC3DP     & \begin{tabular}[c]{@{}l@{}
}Laser profiler \&\\ Depth camera\end{tabular}                         & /                                                                        & Top view  & Filament width                                                                          &                                                                                                                                                            \\ \hline
\cite{Lachmayer2022}     & SC3DP     & Laser profiler                                                                                   & After Nozzle                                                             & Top view  &  \begin{tabular}[c]{@{}l@{}}Filament width, \\ Nozzle height\end{tabular}                                                                          & \begin{tabular}[c]{@{}l@{}}Print speed \& \\ Spray distance\end{tabular}                                                                                   \\ \hline
\cite{WOLFS2024}   & SC3DP     & \begin{tabular}[c]{@{}l@{}}Laser profiler;\\ Near-infrared \\ (NIR) \\ Spectroscopy\end{tabular} & \begin{tabular}[c]{@{}l@{}}Before Nozzle;\\ \\ After Nozzle\end{tabular} & Top view  & \begin{tabular}[c]{@{}l@{}}Filament width\\ \\ Surface water\\ measurement\end{tabular} & \begin{tabular}[c]{@{}l@{}}Flow rate, \\ Cross-sectional area, \\ Material stiffness\end{tabular}                                                          \\ \hline

\cite{Frangez2022_feedback}  & SC3DP     & 4 Depth camera                                                                                   & \begin{tabular}[c]{@{}l@{}}Around \\ the nozzle\end{tabular}             & /         & /                                                                                       & \begin{tabular}[c]{@{}l@{}}C2C/C2M \& \\ Surface classification\end{tabular}                                                                               \\ \hline
\cite{Wolfs2018}   & Extrusion & Laser Dist.                                                                                      & After Nozzle                                                             & Top view  & Nozzle height                                                                           & Filament                                                                                                                                                   \\ \hline
\cite{Kazemian2021} & Extrusion & RGB camera                                                                                       & Before Nozzle                                                            & Top view  & Filament width                                                                          & \begin{tabular}[c]{@{}l@{}}Over-extrusion \& \\ Under-extrusion detection\end{tabular}                                                                     \\ \hline
\cite{Yang2024}    & Extrusion & 2 RGB cameras                                                                                    & \begin{tabular}[c]{@{}l@{}}Before \&\\ After the \\ nozzle\end{tabular}  & Top view  & Filament width                                                                          & \begin{tabular}[c]{@{}l@{}}Pump voltage \&\\ Nozzle travel speed\end{tabular}                                                                              \\ \hline
\cite{Rill-Garcia2022}   & Extrusion & RGB camera                                                                                       & /                                                                        & Side view & Filament height                                                                         & \begin{tabular}[c]{@{}l@{}}Layer to nozzle height,\\ layer thickness,\\ Interlayer line orientation \\ \& curvature,\\ Texture classification\end{tabular} \\ \hline
\cite{Barjuei2022}       & Extrusion & \begin{tabular}[c]{@{}l@{}}Monochrome \\ camera\end{tabular}                                     & Before Nozzle                                                            & Top view  & Filament width                                                                          & \begin{tabular}[c]{@{}l@{}}Motion speed of \\ the nozzle\end{tabular}                                                                                      \\ \hline
\cite{Davtalab2022}      & Extrusion & RGB camera                                                                                       & /                                                                        & Side view & Filament height                                                                         & Defect detection                                                                                                                                           \\ \hline
\end{tabular}
\end{table}

Table~\ref{Table1_online} presents a list of studies that address various aspects of the subject. The table includes papers that employ computer vision sensor techniques for 3DCP, highlighting the preference for computer vision sensors for real-time applications. The AM column, in the table, indicates the printing techniques utilized, while the sensor column specifies the type of sensor employed. The sensor position column shows the position of the sensor relative to the nozzle. The term "after the nozzle" indicates that the sensor is installed posterior to the nozzle in the direction of movement, whereas "before the nozzle" is in the opposite direction. That is to say, "before the nozzle" signifies that the sensor captures what has been deposited from the nozzle immediately, in contrast to "after the nozzle." The column labeled "sensor direction" elaborates on the captured data, specifying whether the sensor captures the top view of the filament (width) or the side view of the filament (height). The "inspections and control" columns state the direct control from the sensor and the end control parameter to be achieved by the author, respectively.

\subsection{Layer-wise}\label{subsec 4.2: Layer-wise}
Various methods for measuring deviations and ensuring quality between different stages of AMC at two levels, namely global and local. Usually, at a global level, a distance (to a point cloud from the previous stage or model) will be assigned to each point. The advantage of this inspection method is its ability to offer a quick overall assessment of the printed object \cite{Nair2022, Mechtcherine2022}.  However, it does not provide detailed local inspection. On the other hand, local inspection offers a detailed analysis of the object's shape and deformation, e.g. filaments’ geometry.

\subsubsection{Global}\label{subsec 4.2.1: Layer-wise-Global}
\begin{figure}[h!]
    \centering
    \begin{tabular}{ccc}
        \includegraphics[width=0.3\textwidth]{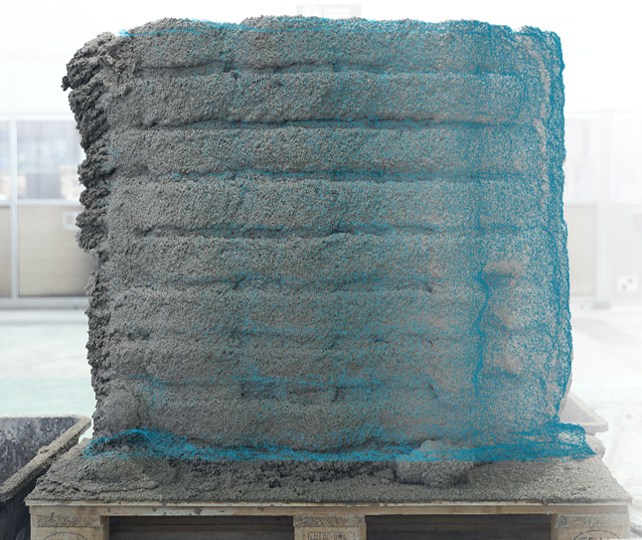}
        & 
        \includegraphics[width=.3\textwidth]{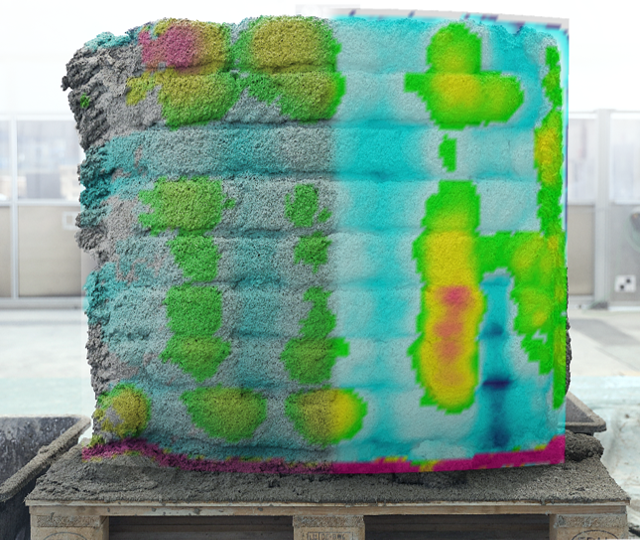}
        &
         \includegraphics[width=.3\textwidth]{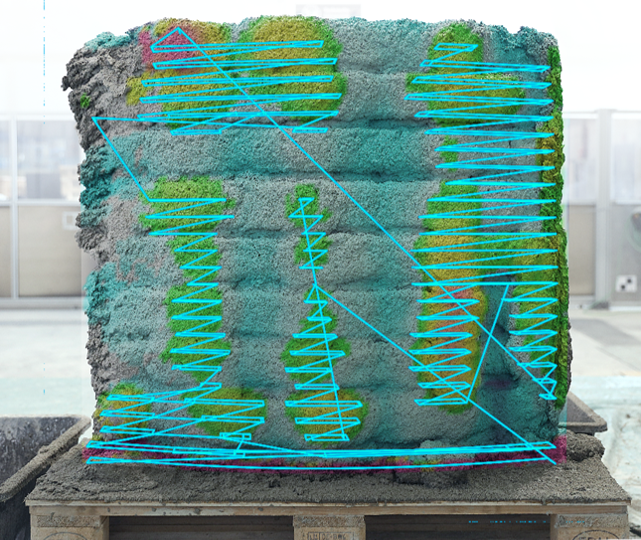}
        \\
        a & b & c \\
    \end{tabular}
    \caption{ QC workflow from data capturing to local path plan generation for defects correction: a) A 3DCP wall with superimposed point cloud. b) A deviation map generated by the C2M method. The map shows regions where too much or less material was applied. c) Automatic local milling path generation for material subtraction based on the deviation map. Images credit: \cite{Hack2022}.
}
    \label{fig: QC-workflow-pathplanning}
\end{figure}

Layer-wise deviation measurement provides holistic information about the current state of the fabricated object, which could be useful for updating the printing parameters to compensate for the deviations in the next step (see Fig.~\ref{fig: QC-workflow-pathplanning}).

A mobile robot equipped with TLS for formative stage scans is employed in \cite{Hack2022}, which facilitated the generation of path plans for subsequent subtractive stages. Deviation measurement methods such as C2C (Cloud-to-Cloud) and C2M (Cloud-to-Model), as well as multiscale model-to-model cloud comparison (M3C2), are usually employed to identify discrepancies between the physical and digital representations of the objects. A practical investigation of these methods is presented in \cite{Mawas2022_Automatic-Inspection} for QC in AMC. To quantify discrepancies of a double-curved 3D-printed wall at both the component and building levels, \cite{Buswell2020} utilized the data from various sensors and employed C2M distance algorithm. For long-term monitoring of 3D-printed walls, \cite{Mendricky2023} utilized an SLS sensor to gain insights into the phenomenon of drying-induced shrinkage.

\subsubsection{Local}\label{subsec 4.2.2: Layer-wise-Local}
\begin{figure}[h!]
    \centering
    \begin{tabular}{cc}
        \includegraphics[width=0.64\textwidth, height=0.25\textheight]{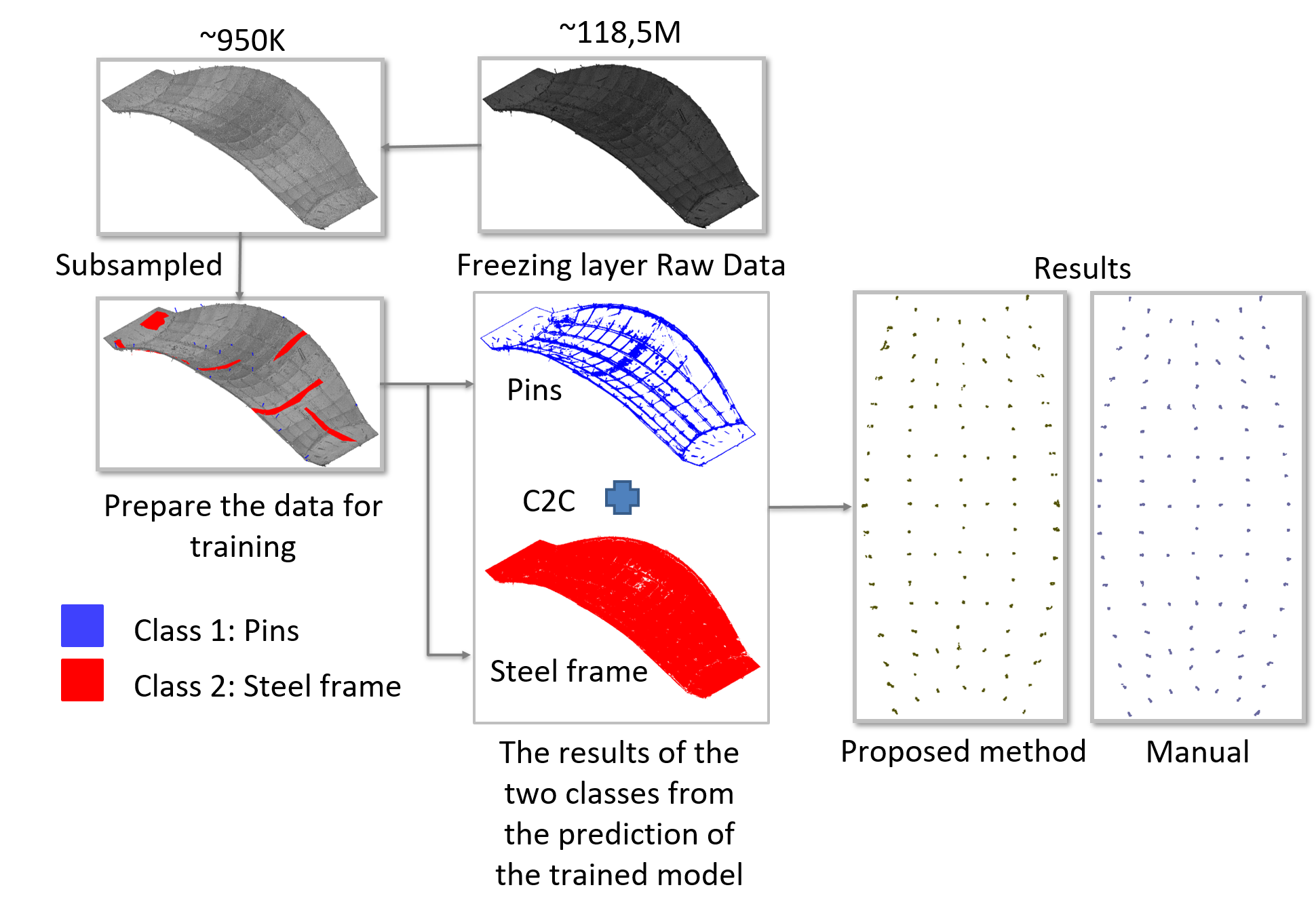}
        & 
        \includegraphics[width=.29\textwidth, height=0.2\textheight]{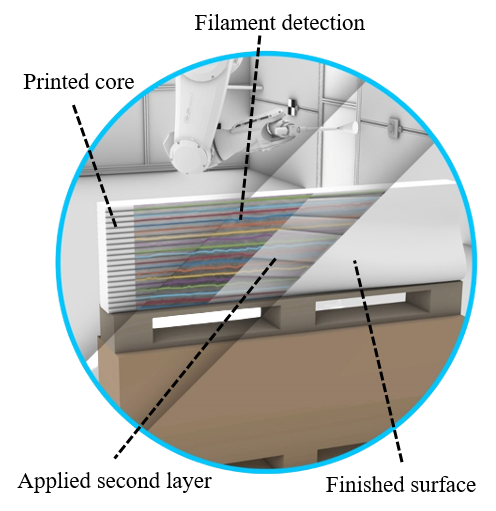}
        \\
        a & b \\
    \end{tabular}
    \caption{Different Local inspection methods for Layer-wise stage. a) Pins detection method from 3D point cloud through a 3D classifier, image modified from: \cite{Rennen2023}. b) Extraction of filaments’ geometry superimposed over the workflow  3D printing  process from printed core to finished surface, image credit: \cite{Mawas2023}.
}
    \label{fig: Layer-wise_Local}
\end{figure}

Local processing is key to getting more insight into small-scale properties, defects and object detection of the fabricated objects. For example, the study conducted in \cite{Rennen2023} employed a semi-automatic method for the extraction of pins (screws) from the 3D point cloud with the objective of determining their position for the knitting process of the fiber winding reinforcement (see Fig. \ref{fig: Layer-wise_Local}a). Moreover, for inspection of the filaments’ geometry, detection of the crack and analysis of the texture of the surface, local inspection should be utilized. For extrusion-based 3DCP, classical image processing techniques, namely edge detection and Hough transform, are reported in \cite{Davtalab2022} to capture filaments’ geometry and identify the defects. The integration of data from a camera and a LiDAR sensor is investigated in \cite{Villacres_2021} to estimate the layer thickness and analyze the deformation of the filaments. For SC3DP, TLS-derived 3D point cloud is converted to 2D images to extract the filaments’ contours \cite{Mawas2023} (see Fig. \ref{fig: Layer-wise_Local}b).

A camera for texture analysis is employed in \cite{Senthilnathan2022}, thereby facilitating the continuous identification of quality issues like cracks on the surface of concrete 3D-printed layers. Supervised and unsupervised machine learning techniques, utilizing depth camera data for material classification and surface quality assessment in SC3DP, are explored in \cite{Frangez2020,Frangez2021}. 
An SLS sensor for micro defect detection in fused deposition modeling (FDM) is used in \cite{Zhao2020}. The author used the region clustering method based on voxel cloud connectivity segmentation (VCCS), introduced in \cite{Papon2013}, to improve the accuracy of surface inspection. Surface defects and cracks detection, using a MobileNet-SSD on images, is employed in \cite{Garfo2020}. Whereas, a Structural Similarity (SSIM) method for assessing filament surface quality during the initial stages is developed in \cite{Fastowicz2018}.

In the context of filament examination, computer vision techniques have the potential to automate quality inspection, thereby reducing the reliance on skilled labor. However, further research is required in SC3DP for filament inspection, with particular attention to the TLS sensor, due to the scarcity of studies on TLS for SC3DP. Additionally, filaments produced by SC3DP are thinner, more deformed, and therefore more challenging to detect than those produced through extrusion-based printing. For micro defect surface inspection, a high-accurate sensor is required, similar to that needed for SLS. In contrast, texture inspection is more suitable with an RGB camera.

\subsection{Preassembly}\label{subsec 4.3: Preassembly}
\subsubsection{Global}\label{subsec 4.3.1: Preassembly-Global}
After surface finishing and edge trimming, the object should be inspected for remaining deviations and defects. Despite conventional concrete components, in AMC, this step is not only for inspection of the component itself but can also be used to send feedback to the design unit to adapt the neighbouring components, if necessary. The C2M distances between as-built point cloud and as-designed model is computed to analyze the quality on a component level \cite{Battaglia2019, Maboudi2020, Buswell2020, Rennen2023}. This methodology also extends to the evaluation of small objects, such as concrete dry joints, for surface smoothness \cite{Baghdadi2023}.

\subsubsection{Local}\label{subsec 4.3.2: Preassembly-Local}
\begin{figure}[h!]
    \centering
    \includegraphics[width=0.8\textwidth, height=0.2\textheight]{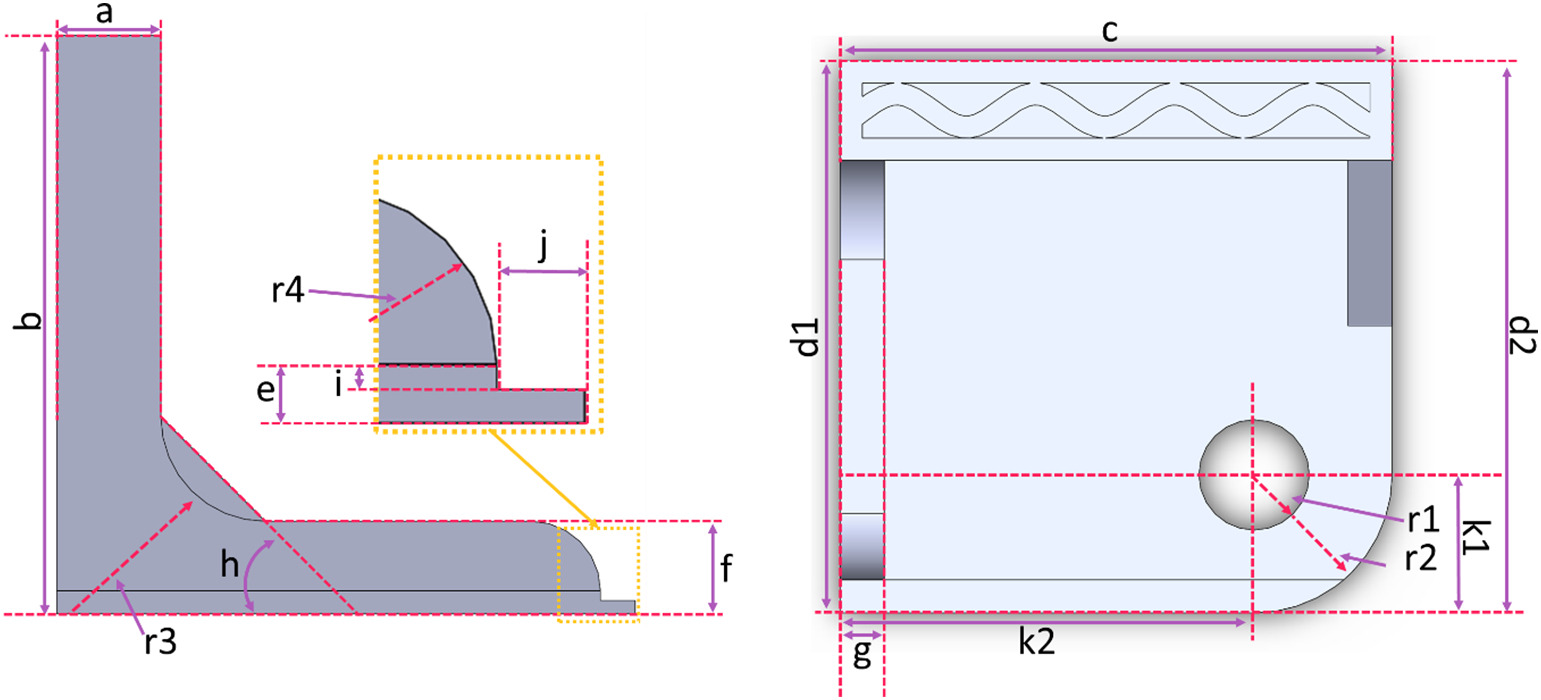}
    \caption{Standard test geometry depicting features and measurement locations, image credit:\cite{BUSWELL2022}.}
    \label{fig: Object_Tolerances_Buswell}
\end{figure}

Beyond global inspection for preassembly, in-depth inspection ensures that 3D-printed objects adhere to their design specifications. Geometric Dimensioning and Tolerancing (GD\&T) is a standardized symbolic language and design tool used in manufacturing and engineering to define and communicate the precise geometric features of parts, ensuring proper assembly, enhanced quality, and cost-effectiveness by clearly specifying size, shape, orientation, location, and tolerance relationships \cite{Cogorno2020}. GD\&T is already standardized in a variety of industries, including automotive \cite{Razak2019}, rail \cite{Jurdeczka2017}, and concrete precast \cite{Wang2017}. GD\&T parameters for 3D concrete printing and inspection standards are also addressed in \cite{XU2020, BUSWELL2022} for maintaining manufacturing tolerances. A standardized test object, designed in \cite{BUSWELL2022} to benchmark the geometric quality of 3DCP (see Fig.~\ref{fig: Object_Tolerances_Buswell}), presents the typical features (dimensions, angles) that are required to be inspected for an object.

A range of approaches and studies have been developed for the evaluation of precast concrete objects, a closely related topic to concrete elements in 3DCP. For instance, an automated dimensional quality technique for precast concrete panels is developed in \cite{Kim2014}. The authors employed a vector sum algorithm to extract edges and corners after plane fitting in a two-dimensional simplification of TLS data. Primitives are detected from point clouds as planes to extract edges and corners \cite{Kim2016, Wang2018}. Point cloud data, of prefabricated components for surface reconstruction using geometric modeling, is processed in \cite{Xu2022}. Further automation is required to achieve quality control with GD\&T metrics within the context of AMC.

Regarding concrete texture, concrete surface quality can be classified into different categories, as exemplified in (\cite{DBV2004, Sichtbetonklassen_Heidelberg, OTTO2024}. Furthermore, the connection between the definition of tolerances in manufacturing and construction \cite{ISO1803, DIN18202} with the standard deviation concept for sensors and definitions in geomantic \cite{DVW2017, Niemeier2017, JCGM_GUM_2023} is a crucial aspect to consider.

\subsection{Assembly}\label{subsec 4.3: Assembly}

Despite using advanced sensors, deviations from the as-designed model can occur, highlighting the need for holistic, multi-stage quality control—before, during, and after assembly- as proposed in \cite{Zhang2020_Quality,Kerekes2022}. For AMC assembly, the QC step is critical to managing discrepancies due to factors such as inaccurate placement of the components, any defects during transportation, environmental changes, and any other unrealized deviations.
\begin{comment}
    \begin{figure}[h!]
    \centering
        \begin{tabular}{cc}
            \multicolumn{2}{c}{\includegraphics[width=0.75\textwidth]{Figures/QC/Assembly/AR_new_correct.png}} \\
            \multicolumn{2}{c}{a}    \\
\includegraphics[width=0.23\textwidth,height=0.13\textheight]{Figures/QC/Assembly/Pavillion_UAV_2.png}        & \includegraphics[width=0.6\textwidth,height=0.17\textheight]{Figures/QC/Assembly/Bridge_The_Gap_Bridge.png}       \\ b & c
    \end{tabular}
    \caption{Different assembly scenarios: a) Construction site with several robots, operators, assembled objects, and highlighted sensors as follows: Vision Trackers in red, Drone in dark gray, TLS in blue, Total station in light gray, AR lenses in light blue looking at demonstrated object in lavender color. b) Results from a pavilion assembly check by a photogrammetry image block captured by a UAV. The points (marked in red) with deviations $>12 mm$, and points (marked in dark blue, values scaled by 100) with deviations $<12 mm$ . Image credit: \cite{Buswell2020}. c) Bridge the gap demonstrator (L.: 5m | H.: 0.8 m | W.: 2.5 m) consists of multi joint connected specimens, image modified from:\cite{AMC-TRR277}.}
    \label{fig:Assembly}
\end{figure}
\end{comment}

\begin{figure}[h!]
    \centering
    \begin{tabular}{c}
        \includegraphics[width=0.7\textwidth]{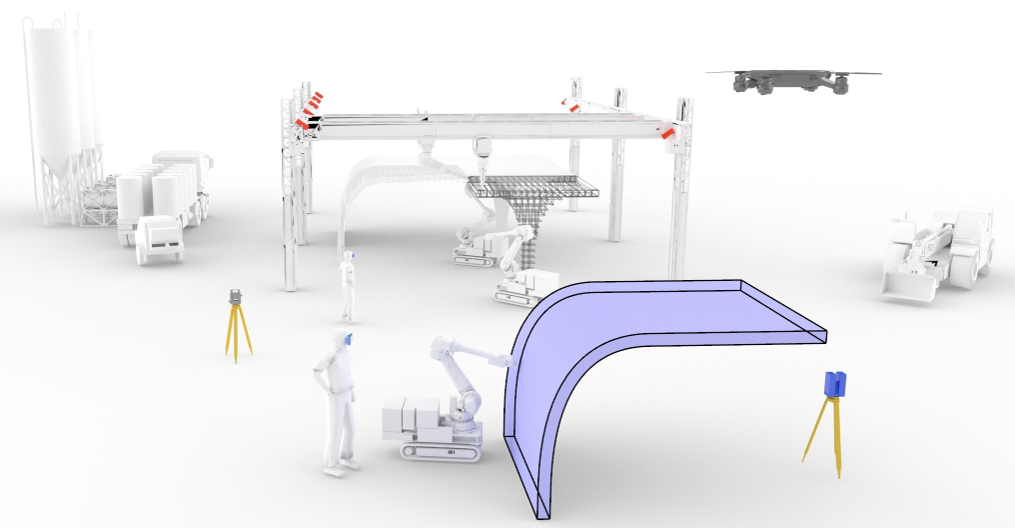} \\
        a \\
    \end{tabular}
    \begin{minipage}[t]{0.49\textwidth}
        \raggedright
        \includegraphics[width=0.55\textwidth]{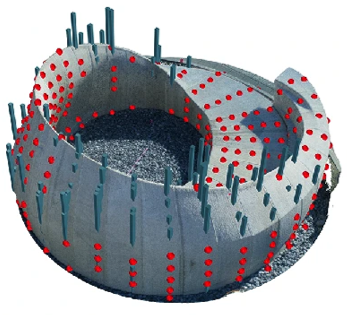} \\
        %\centerline{b}
        \phantom{\hspace{22mm}}b
    \end{minipage}%
    \hfill
    \begin{minipage}[t]{0.49\textwidth}
        \raggedleft
        \includegraphics[width=0.9\textwidth]{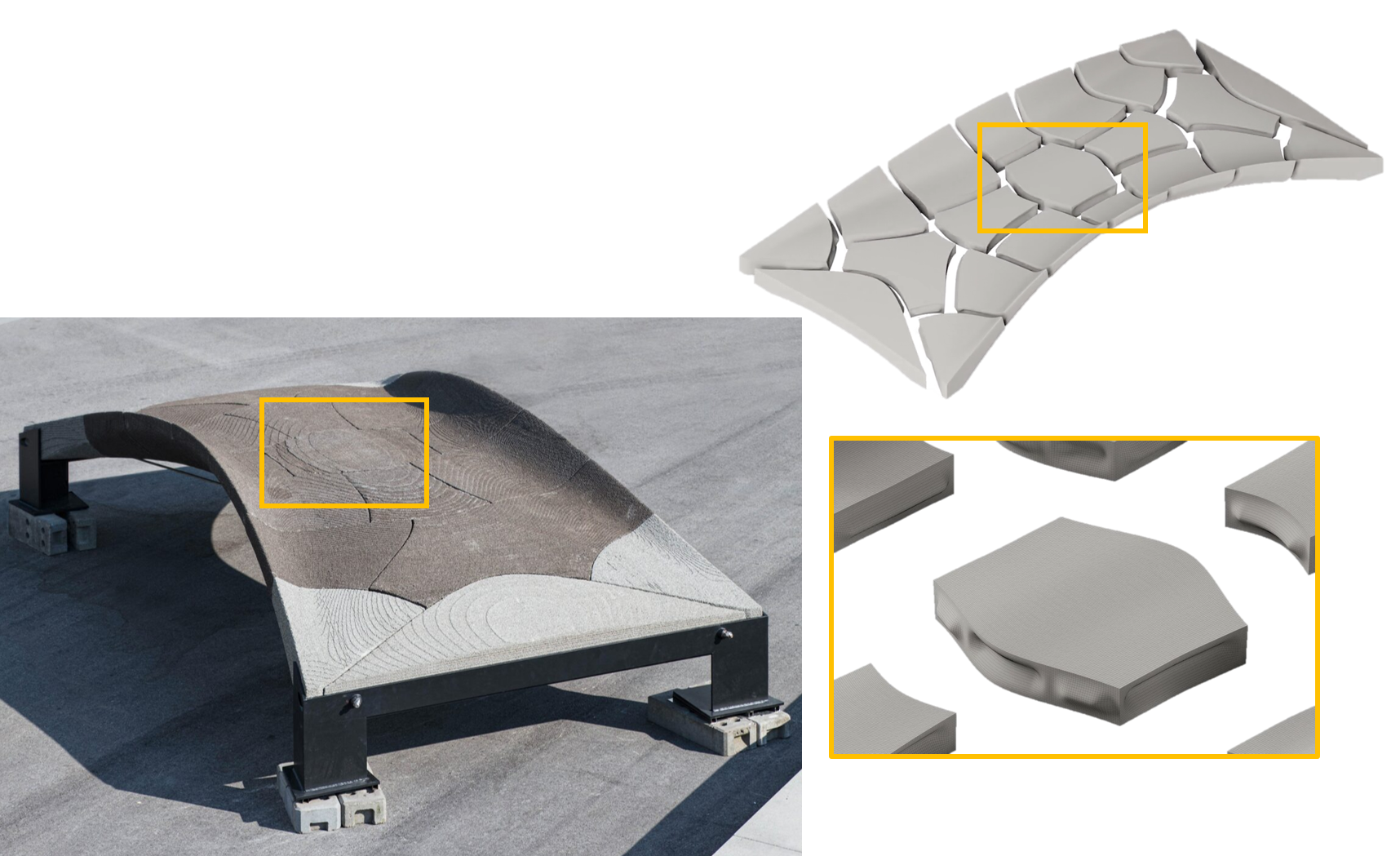} \\
        \centerline{c}
    \end{minipage}
    \caption{Different assembly scenarios: a) Construction site with several robots, operators, assembled objects, and highlighted sensors as follows: Vision Trackers in red, Drone in dark gray, TLS in blue, Total station in light gray, AR lenses in light blue looking at demonstrated object in lavender color. b) Results from a pavilion assembly check by a photogrammetry image block captured by a UAV. The points (marked in red) with deviations $>12 mm$, and points (marked in dark blue, values scaled by 100) with deviations $<12 mm$. Image credit: \cite{Buswell2020}. c) Bridge the gap demonstrator (L.: 5m | H.: 0.8 m | W.: 2.5 m) consists of multi joint connected specimens, image modified from:\cite{AMC-TRR277}.}
    \label{fig:Assembly}
\end{figure}

When joining different parts to form complex structures like bridges \cite{Zhan2021_bridge_assembly,AMC-TRR277} or intricate shapes \cite{Wu2022}, it is essential to ensure that all parts and joints fit correctly according to specified standards (see Fig. \ref{fig:Assembly}c). Achieving these standards within predefined tolerances also depends on selecting and utilizing the appropriate sensors. Various methods, such as C2M comparison \cite{Buswell2020}, help verifying the conformity of the assembled object to its digital replica by using TLS and UAV-based images, under the condition of an accurate registration (see Fig. \ref{fig:Assembly}b). Also, by integrating machine learning algorithms, Augmented/Virtual reality (AR/VR) systems are now capable of identifying structural anomalies, visualization, and filament inspection, for example,  and offering real-time insights \cite{Zimmermann2024}. AR/VR technology allows for the overlay of the digital design model directly onto the physical world, improving the comprehension of project designs and execution on site (see Fig. \ref{fig:Assembly}a). A bidirectional data flow and integration between assembly and structural design for form-finding is reported in \cite{Hack2024}.

QC extends beyond the assembly on the construction site. It also involves ongoing long-term monitoring for continued structure health and safety. A seven-month study comparing captured and digital data to identify deviations and changes is conducted in \cite{Frangez2020}. By doing so, long-term monitoring ensures sustained accuracy and structural integrity. As a result, these different steps of QC provide continuous progress monitoring and accurate BIM updating.

\section{Conclusion and Outlook}\label{sec5: Conclusion}
Several contributions, enhancements, and research have been observed in AMC in a variety of different aspects. However, there are still several areas where improvements can be made, as well as research gaps to be addressed. Additionally, several techniques and methods can be developed and investigated. In this section, we will elaborate on the summary, discussion, and future trends on the topic.

\begin{figure}[b]
    \centering
    \includegraphics[width=0.8\textwidth]{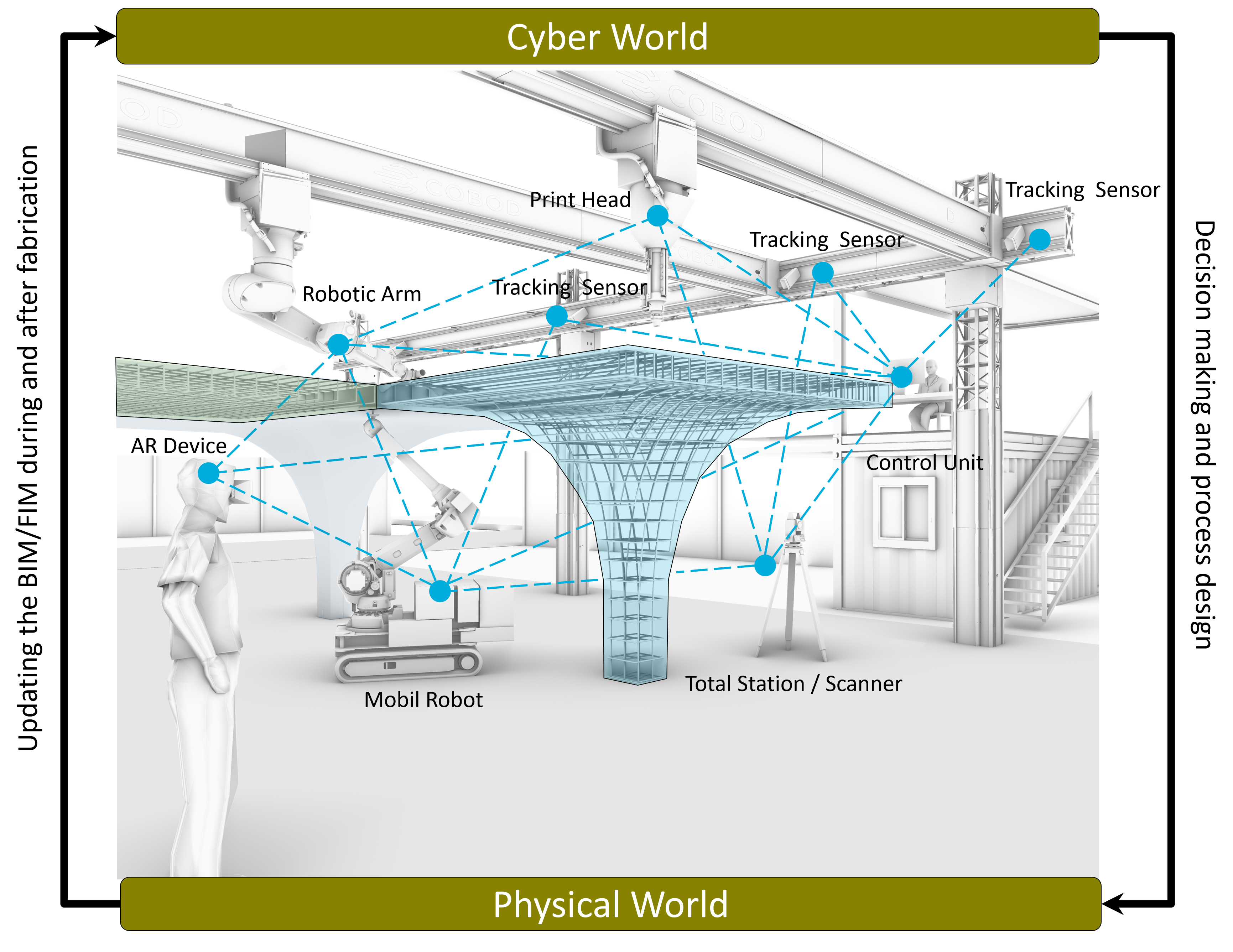}
    \caption{Realization of CPCS for a digital construction site. Image credit: \cite{AMC-TRR277}.}
    \label{fig:Conclusion_CPCS}
\end{figure}

Due to the availability of various printing technologies and the flexibility that 3DCP offers, multiple aspects should be considered for data capturing and QC. These include objects’ properties, environmental conditions, sensors’ specifications,  and the particular requirements of each inspection task. Moreover, site conditions like time limitations, acceptable tolerances, and safety issues should be considered. Therefore, a thorough understanding of each of these aspects is very important to find an optimal setup to achieve the goals of each 3DCP project. Here, data capture planning plays a crucial role in meeting the project-defined quality of the results, adhering to the time constraints and computational limits. Another very important aspect is establishing a common coordinate system to bring the existing design information and newly acquired data into a common spatial and temporal frame for further analysis and decision-making. Particular attention should be devoted to effective data management to ensure that the gathered information is well-structured, informative,  and readily usable for further visualization, systematic reporting, efficient querying, and seamless communication.
\\~\\
\contourlength{0.25pt}
\contour{black}{Sensors}: In selecting a proper sensor for data capture and QC there is no \textit{“one fit all”} solution. It is not only related to the specifications of the available sensors, but it also depends on the object properties, environmental conditions, and project demands. One important outcome of our literature study is that as much as \textbf{the sensor could be integrated into the fabrication system}, post-processing of the data, especially the registration of the captured data to the reference coordinate system would be less exhaustive. Moreover, since this is a multi-criteria decision, rule-based selection could be a wise choice. While rule-based solutions are white-box processes, they are easily interpretable. Hence, even if none of the already existing sensors in a project pass all the selection rules, it is clear which parameters should be changed or which sensor should be added. For example, if, for a specific task, a photogrammetry solution passes all the conditions except the environmental conditions (for example, lighting condition in this case), the data capture and QC  team might find a solution for that (for example, using artificial light). It is worth mentioning that the availability of more affordable sensors enables their utilization by mid-sized and small enterprises. The current use of TLS for 3DCP remains limited compared to other sensors, indicating a significant opportunity for further research and development. Developing multi-sensor systems or sensor fusion strategies can be a topic for future research for further and better data understanding and analysis.
\\~\\ 
\contour{black}{Data capture planning}: For each type of the selected sensors, different approaches for capture planning exist that are discussed in sec. \ref{sec3: Data Capturing}. Since \textbf{as-designed models of the object are already existing in 3DCP}, the capture planning could be more effective and can be performed before printing in simulated environments. It is important to emphasize that the data capture planning should be adaptive enough to incorporate necessary parameters, namely, time, LOA, LOD, and LOC for each task. The possibility of reaching the computed position and orientation of each data capture pose should also be considered. This is at least a two-fold problem. One aspect is \textbf{finding a solution to reach the computed sensor pose}, especially for large objects. Another aspect is the practical accuracy of reaching such sensor poses. Mounting the sensor on the printing robot could be a solution to handle both aspects. When the object is too big, it is very likely that the manufacturing site is also big and GNSS signals are available. In this case, agile platforms like UAVs can be a solution to reach the predefined poses for data capturing with lightweight sensors. Moreover, employing backpack data capturing systems, and cyberdogs could also be game-changing for data capturing in construction sites with dynamic environments and a lot of clutters. Additionally, virtual reality (VR) and augmented reality (AR) represent a potential platform for data capture, QC, and data visualization \cite{Khoshelham2019, Hubner2020}.
\\~\\ 
\contour{black}{Registration}: Although there are well-established registration methods for rigid objects (cf. sec. \ref{subsec3.3: Registration}), special considerations in AMC need to be regarded. AMC necessitates the stage-wise QC while during manufacturing the object is not rigid and also material addition changes the shape of the object in the previous state. Moreover, \textbf{the need for higher automation and limited time during printing demands customized and well-thought data capturing strategies}. Within the limited time between manufacturing stages, efficient and reliable data capture, registration, QC, and feedback to the design unit,  allows for possible modifications, if necessary.  In this regard, two future works -- among others -- include point cloud to model coregistration and integration of sensing devices in printing setup. While the former is an active research topic, the latter should be investigated more while considering the printing requirements and conditions. For example, if TLS is being employed, mounting the TLS on the robotic arm could be a solution to the registration problem. This necessitates a carefully managed calibration of the TLS to the reference coordinate system, e.g. fabrication coordinate system. To ensure precise registration of the diverse sensor data into a unified coordinate system, it is possible to employ an accurate tracking system. 
\\~\\ 
\contour{black}{QC}: The quality of the printed structure in its hardened state depends on a combination of factors, including the properties of the fresh mixture, deposition parameters, printer technology, and surface finishing tools. These factors span both the printing process and the post-printing phase, highlighting the need for \textbf{QC at various stages of manufacturing}. Data capturing for QC for 3DCP has been performed using multiple sensors, including 1D, 2D, 2.5D, and 3D sensors. Computer vision methodologies have demonstrated notable efficacy and have motivated research in this area. Additionally, computer vision techniques have the potential to automate quality inspection, thereby reducing the reliance on skilled labor. However, further research is required in SC3DP for filament inspection. 3D point cloud processing using 3D deep learning methods is an interesting topic to investigate, such as filament extraction and segmentation. Moreover, further automation is required to achieve QC with \textbf{geometric dimensioning and tolerancing} metrics within the context of AMC. Additionally, the relationship between this and various AMC standards, as well as the surface quality of concrete, is worthy of note. Investigation of shrinkage effects as well as \textbf{long-term monitoring for 3DCP} is crucial to perform, considering the variety of different sensors.
\\~\\ 
\contour{black} {Outlook regarding data integration, visualization, querying, and reporting}:
The establishment of new architectural infrastructure is imperative for the integration of bi-directional communication between the cyber and physical worlds in real-time. This infrastructure must facilitate the integration and synchronization of diverse data, platforms, machines, sensors, and humans into a unified system and coordinate system \cite{Sawhney2020}. The construction of such a platform necessitates the implementation of a \textbf{cyber-physical construction system (CPCS)}, which integrates disparate actuators and sensors, thereby facilitating a bidirectional data flow between the physical and cyber domains. Furthermore, as illustrated in Fig. \ref{fig:Conclusion_CPCS}, the CPCS paradigm shows a transition from pure automation to \textbf{human-robot collaboration}, thereby enabling \textbf{human-centric} and creative processes within the context of \textbf{Industry 5.0} \cite{Xu2021, Leng2022}. 

Effective and efficient integration of QC results, geometric dimensioning and tolerancing (GD\&T), and other relevant aspects into the well-established data center requires further exploration and investigation. Other important aspects of data sharing, e.g. communication technologies and protocols (such as 5G, 6G, etc), standardization, validity, cyber security, and methods for big-data handling, to name a few, are also the current topics of research. These essential aspects will affect and lead to more effective integration between business models and technical solutions to enhance data sharing in the construction industry \cite{Wang2023, Anumba2021}.

\begin{comment}
    \begin{figure}[h!]
    \centering
    \begin{tabular}{cc}
        \includegraphics[width=0.48\textwidth, height=0.23\textheight]{Figures/Conclusion/CPCS_2.png}
        & 
        \includegraphics[width=.48\textwidth, height=0.3\textheight]{Figures/Conclusion/CPCS_C06_2.png}
        \\
        a & b \\
    \end{tabular}
    \caption{Realization of Industry 4.0 for construction within the CPCS concept.  a)  Conceptual illustration of the Construction 4.0 framework. Image credit: \cite{Sawhney2020}, b) Realization of CPCS for a digital construction site. Image credit: \cite{AMC-TRR277}.}
    \label{fig:Conclusion_CPCS}
\end{figure}
\end{comment}

Advancements in image and point cloud deep learning models have been observed in recent years, demonstrating generic and multitasking capabilities, such as segmentation \cite{Kirillov2023_SAM}, tracking and object detection \cite{Lazarow2024, ultralytics}. The integration of such models, along with the use of VR/AR lenses, facilitates direct visualization of the processes and assists the operators in the execution of QC tasks for AMC \cite{Mitterberger2023}. Moreover, the recent introduction of various \textbf{Large Language Models (LLMs)} opened a new dimension for text and speech analysis. For example, LLMs have the potential to increase the efficiency for the generation of reports and technical documents, as well as design proposals, through project-specific queries \cite{Rane2023, Gao2024}. Hence, the integration of a \textbf{Vision Language Model (VLM)}, derived from the combination of both language and vision models, has the potential to enhance and improve the interaction between machines and humans.

Recently, the efficacy of domain-specific knowledge surpassed generic LMs \cite{Jiang2024}. Furthermore, two distinct methodologies are compared in \cite{Lee2024}, namely: the employment of a Retrieval-Augmented Generation (RAG) model and the fine-tuning of an LLM with a generic-LLM (GPT-4) to facilitate the management and understanding of construction safety standards which include both textual and visual elements. Similarly, Visual Question Answering (VQA) approach for the construction industry is proposed in \cite{Yang2024_Transformer}, with the objective of addressing the unique requirements of on-site management. The development of such \textbf{AMC-specific Large Vision Language Model (LVLM)} models for construction purposes customized to align with AMC domain-specific knowledge could be an interesting research direction. Augmenting time-stamped 2D and 3D information from QC in such models for further querying and reporting is a leap toward automated QC in AMC. 

\bmhead{Acknowledgements}
This research is funded by the Deutsche Forschungsgemeinschaft (DFG, German Research Foundation) – TRR 277/2 2024 – Project number 414265976. The authors thank the DFG for the support within the CRC / Transregio 277 - Additive Manufacturing in Construction (Project C06: Integration of Additive Manufacturing into a Cyber-Physical Construction System). 

We would like to express our profound gratitude to Abtin Baghdadi, Robin Dörrie, and Sven Jonischkies from ITE TU Braunschweig for providing the as-designed models, which were instrumental in creating the figures.
%TUBS Acknowledgements for open source paper.

\begin{comment}
\section*{Declarations}
AI tools such as DeepL Write and Grammarly have been used to improve English writing and rectify grammatical errors, thus improving the readability of this paper.    
\end{comment}

%Some journals require declarations to be submitted in a standardised format. Please check the Instructions for Authors of the journal to which you are submitting to see if you need to complete this section. If yes, your manuscript must contain the following sections under the heading `Declarations':

\bibliography{sn-article}% common bib file

\end{document}